\documentclass[review,5p,twocolumn]{elsarticle}
\usepackage{amsmath,amsfonts}
\usepackage{algorithmic}
\usepackage{algorithm}
\usepackage{array}
\usepackage{textcomp}
\usepackage{stfloats}
\usepackage{url}
\usepackage{verbatim}
\usepackage{graphicx}
\usepackage{xcolor}
\usepackage{nicematrix}
\usepackage{emptypage}
\usepackage{fancyhdr}
\usepackage{graphicx}
\usepackage{float}
\usepackage{subcaption}
\usepackage{caption}
\usepackage{braket}
\usepackage{pifont}
\usepackage{amssymb}
\usepackage{multirow}
\usepackage{tabularray}
\usepackage{hyperref}
\usepackage{mathtools}
\usepackage{lineno}
\usepackage{booktabs}
\modulolinenumbers[5]

\begin{document}

\begin{frontmatter}

\title{Hybrid Quantum-inspired Resnet and Densenet for Pattern Recognition}
\tnotetext[mytitlenote]{corresponding author}

%% Group authors per affiliation:
%\author{Andi Chen$^{a,b}$, Hua-Lei Yin$^{c,*}$, Zeng-Bing Chen$^{d,*}$, Shengjun Wu$^{a,b,d,*}$}
%\address{Radarweg 29, Amsterdam}
%\fntext[myfootnote]{Since 1880.}
%\author[mymainaddress,mysecondaryaddress]{Andi Chen}
\author{Andi Chen$^{a,b}$}
\ead{andynju1999@gmail.com}

\author{Hua-Lei Yin$^{c,\star}$}
%\cortext[mycorrespondingauthor]{Co-corresponding author}
\ead{hlyin@ruc.edu.cn}

\author{Zeng-Bing Chen$^{b,\star}$}
%\cortext[mycorrespondingauthor]{Co-corresponding author}
\ead{zbchen@nju.edu.cn}

\author{Shengjun Wu$^{a,b,\star}$}
%\cortext[mycorrespondingauthor]{Co-corresponding authors}
\ead{sjwu@nju.edu.cn}
%% or include affiliations in footnotes:
%\author[mymainaddress,mysecondaryaddress]{Elsevier Inc}
%\ead[url]{www.elsevier.com}

%\author[mysecondaryaddress]{Global Customer Service\corref{mycorrespondingauthor}}
%\cortext[mycorrespondingauthor]{Corresponding author}
%\ead{support@elsevier.com}
\address[mymainaddress]{Institute for Brain Sciences and Kuang Yaming Honors School, Nanjing University, Nanjing 210023, China}
\address[mysecondaddress]{National Laboratory of Solid State Microstructures and School of Physics, Collaborative Innovation Center of Advanced Microstructures, Nanjing University, Nanjing 210093, China}
\address[mythirdaddress]{School of Physics and Beijing Key Laboratory of Opto-Electronic Functional Materials and Micro-Nano Devices, Key Laboratory of Quantum State Construction and Manipulation(Ministry of Education), Renmin University of China, Beijing 100872, China}

\begin{abstract}
In this paper, we propose two hybrid quantum-inspired neural networks with adaptive residual and dense connections respectively for pattern recognition. We explain the frameworks of the symmetrical circuit models in the quantum-inspired layers in our hybrid models. We also illustrate the potential superiority of our hybrid models to prevent gradient explosion owing to the sine and cosine functions in the quantum-inspired layers. Groups of numerical experiments on generalization power showcase that our hybrid models are comparable to the pure classical models with different noisy datasets utilized. Furthermore, the comparison between our hybrid models and a state-of-the-art hybrid quantum-classical convolutional network demonstrates 3\%-4\% higher accuracy of our hybrid densely-connected model than the hybrid quantum-classical network. \textcolor{black}{Additionally, compared with other two hybrid quantum-inspired residual networks, our hybrid models showcase a little higher accuracy on image datasets with asymmetrical noises.} Simultaneously, in terms of groups of robustness experiments, the outcomes demonstrate that our two hybrid models outperform pure classical models notably in resistance to adversarial parameter attacks with various asymmetrical noises. They also indicate the slight superiority of our densely-connected hybrid model over the hybrid quantum-classical network to both symmetrical and asymmetrical attacks. \textcolor{black}{Meanwhile, the accuracy of our two hybrid models is a little bit higher than that of the two hybrid quantum-inspired residual networks.} In addition, an ablation study indicate that the recognition accuracy of our two hybrid models is 2\%-3\% higher than that of the traditional quantum-inspired neural network without residual or dense connection. Eventually, we discuss the application scenarios of our hybrid models by analyzing their computational complexity.
\end{abstract}


\end{frontmatter}

%\linenumbers

\section{Introduction}
As the cornerstone of AI technology, deep neural network algorithms have played a pivotal role over the past few decades \cite{1,2,3,4,5,6,7}. Typical models incorporate deep residual and dense networks (Resnet and Densenet) in image recognition, Large Language Models (LLM) in natural language processing, and Sora model in multi-modal generations \cite{8,9,10,11}. Nevertheless, due to the limited generalization power and robustness of classical neural networks, novel neural networks with stronger generalization power under assorted environments and robustness to adversarial attacks over pure classical models is tremendously expected nowadays \cite{12,13,14,15,16}. Simultaneously, quantum-inspired neural networks, which obey the laws of quantum computing and deep learning, appear with their prominent performances on certain circumstances \cite{17,18,19}. Innovative frameworks incorporate quantum-inspired stochastic walks and variational circuit models, etc \cite{19,20,21,22}. While stochastic walks incorporate the solution and evolution of Hamiltonian, variational circuit models involve continuous multiplication of unitary matrices. What's more, to integrate the upsides of pure classical neural networks with that of quantum-inspired networks, hybrid quantum-inspired neural networks have been explored in some studies \cite{23,24,25,26}. Amalgamated with classical convolutional layers, they demonstrate remarkable performance on image detection and other tasks \cite{49,50,51}. Comprised of the pure classical part and the quantum-inspired part, they are applicable to more types of problems \cite{25,26}. 

\textcolor{black}{Nevertheless, despite a few works of hybrid quantum-inspired models, the experimental results of them showcase that those hybrid quantum-inspired networks only either show higher accuracy in pattern classification problems, or possess stronger generalization ability in different scenarios, or are more robust than traditional classical or quantum neural networks. Additionally, models with distinguished robustness possess lower accuracy or weaker generalization ability than classical ones \cite{23,25,26,49,50,51}. There has been no hybrid quantum-inspired models possessing both high accuracy and enhanced generalization capability as well as robustness to adversarial attacks. Meanwhile, while several quantum neural networks with circuit models demonstrate unique robustness to noise attacks, numerous classical residual and dense neural networks showcase exceptional accuracy and generalization power \cite{2,3,8,9,21,22}. What's more, a common problem on quantum neural networks is barren plateau, which could be avoided by the addition of residual frameworks \cite{23}. Concurrently, there is a lack of hybrid quantum-inspired models leveraging the amalgamation of circuit models and residual or dense connection \cite{8,9,27}. Hence, to facilitate the universality and comprehensiveness of hybrid quantum-inspired models under complex environments,} we firstly devise the hybrid $\textbf{Q}$uantum-inspired $\textbf{R}$esidual $\textbf{F}$eedforward $\textbf{N}$eural $\textbf{N}$etwork (QRFNN) and hybrid $\textbf{Q}$uantum-inspired  $\textbf{D}$ense $\textbf{F}$eedforward $\textbf{N}$eural $\textbf{N}$etwork (QDFNN). We also devise hybrid $\textbf{Q}$uantum-inspired $\textbf{R}$esidual $\textbf{C}$onvolutional $\textbf{N}$eural $\textbf{N}$etworks (QRCNN) and hybrid $\textbf{Q}$uantum-inspired $\textbf{D}$ense $\textbf{C}$onvolutional $\textbf{N}$eural $\textbf{N}$etworks (QDCNN) for image problems. We apply our four hybrid models on pattern classification problems. A $\textbf{H}$ybrid $\textbf{Q}$uantum-classical convolutional $\textbf{N}$etwork (HQNet), pure classical $\textbf{M}$ulti-$\textbf{L}$ayer $\textbf{P}$erceptrons (MLPs), $\textbf{C}$onvolutional $\textbf{N}$eural $\textbf{N}$etworks (CNNs) and two hybrid $\textbf{Q}$uantum-inspired $\textbf{R}$esidual neural $\textbf{N}$etworks (ResQNet1 and ResQNet2) with detailed structures serve as state-of-the-arts for model comparison. Hence, in the paper, we: \\
\begin{itemize}
\item design and analyze QRFNN and QDFNN for recognition of iris data \cite{28} under noisy and noiseless environments;
\item design and analyze QRCNN and QDCNN for classification of MNIST, FASHIONMNIST and CIFAR image datasets \cite{28} under noisy and noiseless environments;
\item conduct assorted numerical experiments to compare our hybrid models with a state-of-the-art hybrid quantum-classical convolutional network, pure classical models and two hybrid quantum-inspired residual networks;
\item implement an ablation study to compare our hybrid models with traditional quantum-inspired neural networks without residual or dense connection on pattern recognition;
\item discuss the advantages and illustrate the application scenarios of our hybrid models owing to their frameworks and their computational complexities.
\end{itemize}
\textcolor{black}{The contributions of this paper are that:}
\begin{itemize}
	\item \textcolor{black}{we devise hybrid quantum-inspired neural networks with proposed adaptive residual or dense connection. We also design symmetrical "V" circuits with $N-2$ unitary sparse matrices forming one quantum-inspired layer as the residual or dense block, where $N$ is the dimension of the Hilbert space in the quantum-inspired layers;} 
	\item \textcolor{black}{our densely-connected hybrid models marginally outperform the state-of-the-art HQNet on generalization power when the datasets contain symmetrical or asymmetrical noises;}
	\item \textcolor{black}{our two hybrid models demonstrate a little bit higher accuracy than other two state-of-the-art hybrid quantum-inspired residual networks to asymmetrical noise attacks or on image classification when the datasets contain asymmetrical noises;}
	\item \textcolor{black}{our hybrid models surpass classical MLPs and CNNs a lot on robustness to asymmetrical noise attacks;}
	\item \textcolor{black}{by analyzing the loss functions and parameter gradients of classical and our hybrid models, the reason why our hybrid models could resist asymmetrical noise attacks lies in their capability to prevent gradient explosion.}
\end{itemize}

Next, in section 2, we review the corresponding works of classical residual and dense frameworks and quantum-inspired neural networks with circuit models. In section 3, we introduce some basic concepts of circuit models, residual and dense structures \cite{8,9,27}. As for section 4, we explain our hybrid models including the layer details and parameter learning details. Following it, numerous experiments are implemented in section 5, which involve comparisons between our hybrid models and pure classical models, the hybrid quantum-classical models and other two hybrid quantum-inspired residual networks. We also conduct an ablation study incorporating comparisons between our hybrid models and the quantum-inspired neural network without residual or dense connection in section 5. Finally, we illustrate the application scenarios and some future works of our hybrid models in section 6.

\section{Related Work}
\subsection{Residual and dense frameworks} 
The original deep neural network for pattern recognition, Alexnet, was proposed in 2012 \cite{29}. In spite of the outstanding performance of deep networks in complicated problems \cite{30,31,32,33,34,35,36,37}, experimental findings had also underscored the trouble in training deep neural networks without residual or dense connection, such as feature disappearance \cite{1,2,8,9}. Consequently, Resnet and Densenet, which were firstly introduced in 2016, showed outstanding performance in image classification \cite{8,9}. The residual and dense frameworks imply the superiority of the element-wise addition mechanism and the feature map concatenation mechanism separately. Additionally, both of the two approaches show a great capacity to mitigate feature vanishing and overcome gradient vanishing \cite{2,8,9}. Large-scale models such as YOLO in autonomous driving and attention-based Transformer in text translation, draw inspiration from these seminal works \cite{31,38}. Therefore, the residual and dense connections may facilitate feature propagation and enhance the universality of the quantum-inspired part of our hybrid models as well. Nevertheless, if there is some noise in the datasets, the original residual and dense frameworks may also possess the potential to transmit the noise information to the output layer while transferring the feature of the data itself, thereby interfering with recognition accuracy \cite{1}. Therefore, different from the original residual and dense architectures, we propose two adaptive residual and dense connections with adaptive parameters $\lambda$ that can update themselves by back propagation to reduce the impacts of noise. And we embed the adaptive residual and dense architectures into the quantum-inspired layers of our hybrid models.  
\subsection{Quantum-inspired neural networks with circuit models}
Quantum-inspired neural networks, a classical algorithm essentially, are impacted by some quantum laws \cite{21,22}. A typical model is circuit model \cite{22,39}. In the authentic quantum circuit system, classical data is encoded into a quantum state. And the data feature is acquired via evolutionary processes and stored in the parameters in the unitary gates of the circuit models. And the design of the quantum-inspired part of our hybrid models draw on the frameworks of the circuit models in authentic quantum system \cite{22}. Moreover, our quantum-inspired layers only follows some laws of circuit models, therefore, our hybrid models require more improvements to be deployed on authentic quantum computers. However, in previous studies of circuit models, the architectures of the circuits are usually asymmetrical, which may result in information loss \cite{22}. Moreover, the unitary gates utilized incorporate ${\rm R}_{\rm Y},{\rm R}_{\rm Z}$, Hadamard and CNOT gates, some of which contain no parameter and could not assist feature learning \cite{25,26,39}. Hence, to enable our models to fully extract the feature, each gate in the quantum-inspired part contains a parameter. Moreover, one novelty lies in the proposal of the symmetrical "V" shape in each quantum-inspired layer of our hybrid models in Fig.3, which may also help feature storage \cite{19,40}. \textcolor{black}{In ref. \cite{nie}, the authors compare the accuracy of symmetrical and asymmetrical circuits for data recognition. For each category of the data, symmetrical circuits possess enough parameters to learn their features. In contrast, asymmetrical circuits may cause imbalance in feature learning, which means they are more efficient and accurate in extracting features from certain categories of a dataset than from other categories of the same dataset\cite{2,nie}. In ref. \cite{40}, experiments have showcased prominent accuracy of the "V" shape network combined with residual connections in image classification problems as well.} In addition, traditional quantum-inspired networks may not possess very outstanding learning capability \cite{21}, thus another novelty lies in the combination of the adaptive residual and dense connections and the symmetrical circuit models.

\section{Preliminary}
\subsection{Circuit models}
Analogous to classical bit in classical computers, a qubit, the unit in quantum-inspired algorithms, is described as a 2-dimensional complex vector in Hilbert space \cite{22}. It can be described according to the superposition rules:
\begin{equation}
	\label{deqn_ex1a}
	\ket{\psi} = a \ket{0} + b \ket{1}, \enspace |a|^2 + |b|^2 = 1 \; (a,b \in \mathbb{C}).
\end{equation}

In our hybrid frameworks, the evolution process stands for the manipulation to the quantum states with continuous unitary operations:
\begin{equation}
	\label{deqn_ex1a}
	\ket{\psi_m} = U_{m} \ket{\psi_{m-1}}\quad(m \in \mathbb{N}^+),
\end{equation}
where $U_m$ represents the $(m+1)^{th}$ unitary gate in the $m^{th}$ evolution. And $\rm{\ket{\psi_0}}$ means the initial state.The gate $T_{i}(\theta^{l-1}_r)$, frequently used in the paper, can be written in the computational basis as the following matrix
\begin{equation}
	\begin{scriptsize}
		\label{deqn_ex1a}	
		\begin{pNiceMatrix}[first-col,last-row]
			&   1  & 0    &\cdots &\cdots & \cdots & \cdots &\cdots & 0    \\
			&   0  & 1    &      &           &            &      &   &  \vdots    \\
			& \vdots &      &\ddots&           &            &      &   & \vdots     \\
			i   &\vdots&      &      &\cos({\mathrm\theta}^{l-1}_r)&{\rm -sin}(\theta^{l-1}_r)&      &   & \vdots     \\
			i+1  &\vdots	    &      &      &\sin({\rm\theta}^{l-1}_r)&\cos({\rm\theta}^{l-1}_r) &    &   &\vdots\\
			&\vdots	    &      &      &           &            &\ddots&   & \vdots     \\
			&\vdots	    &      &      &   & &      & 1 & 0    \\
			&    0&\cdots&\cdots &\cdots&\cdots &\cdots& 0 & 1    \\
			&     &      &      &   i       &  i+1       &      &   &      \\
		\end{pNiceMatrix}
		.
	\end{scriptsize}
\end{equation}

As for Eq.(3), $i$ denotes the row index of the term ${\rm -sin}({\theta}^{l-1}_r)$, also the column index of the term $\sin({\theta}^{l-1}_r)$ $(i \in \mathbb{N}^+)$. And $\theta^{l-1}_r$ means the $(r+1)^{th}$ parameter in the $l^{th}$ layer $(r \in \mathbb{N}, l \in \mathbb{N}^+)$. And $r$ has relations with $i$, shown in Eq.(4).
\subsection{Resnet and Densenet learning}
Realizing the possible impacts from the noise, we propose linear addition mechanism with adaptive parameters \cite{2,8}. In terms of the residual framework, Fig.1 shows its partial structure. $H_h(x) = R_{h-1}(H_{h-1}(x)) \oplus \lambda_{h-1} H_{h-1}(x)$ $(h \in \mathbb{N}^+)$, where $\oplus$ means linear element-wise addition, $R_{h-1}$ is the residual mapping of the $h^{th}$ layer. And $\lambda_{h-1}$ is the adaptive parameter of $H_{h-1}$ in the $h^{th}$ layer. And the dense framework in this paper also emanates from the element-wise addition mechanism of all the preceding layers in Fig.2, where $G_a(x) = D_{a-1}(G_{a-1}(x)) \oplus \lambda_{a,0} x \oplus \sum\limits^{a-1}_{b=1} \lambda_{{a,b}}G_b(x)$$(a,b \in \mathbb{N}^+)$ for more layers. $D_{a-1}$ refers to dense mappings of the $a^{th}$ layer. $\lambda_{a,0}$ and $\lambda_{a,b}$ are the adaptive parameters of $x$ and $G_b(x)$ respectively.
\begin{figure}[!t]
	\centering
	\includegraphics[width=3.5in]{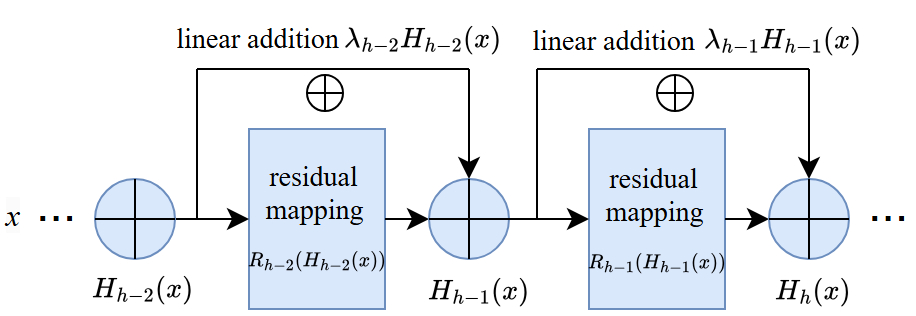}
	\caption{Adaptive residual connection}
	\label{Resnet}
\end{figure}

\begin{figure}[!t]
	\centering
	\includegraphics[width=3.5in]{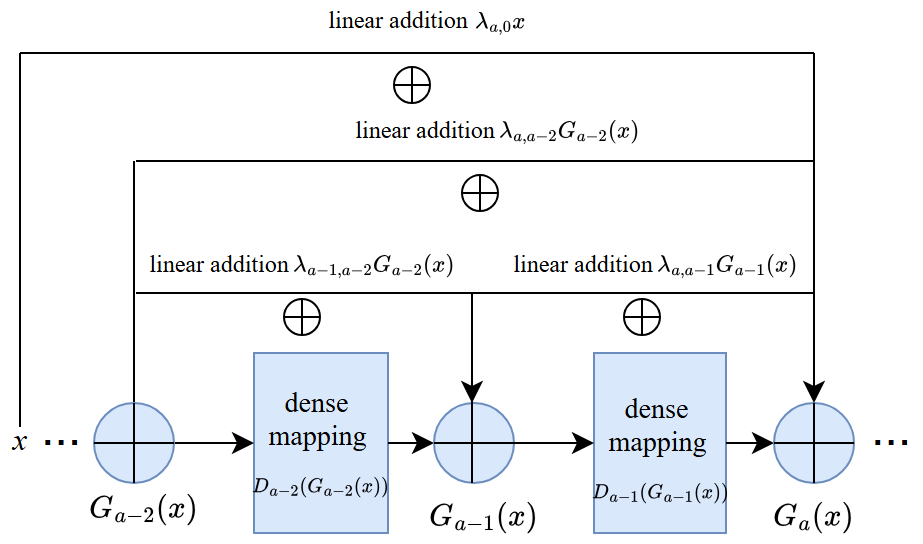}
	\caption{Adaptive dense connection}
	\label{Densenet}
\end{figure}

\begin{figure*}[!t]
	\centering
	\subfloat[\centering]{\includegraphics[width=6.7in]{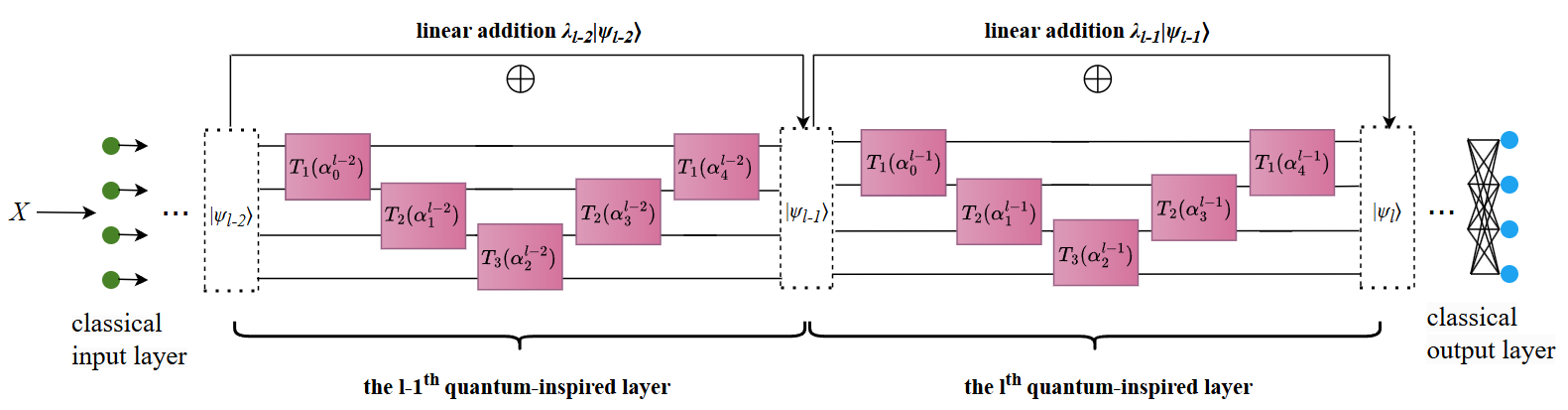}%
		\label{QRFNN}
	}
	\hfil
	\subfloat[\centering]{\includegraphics[width=6.7in]{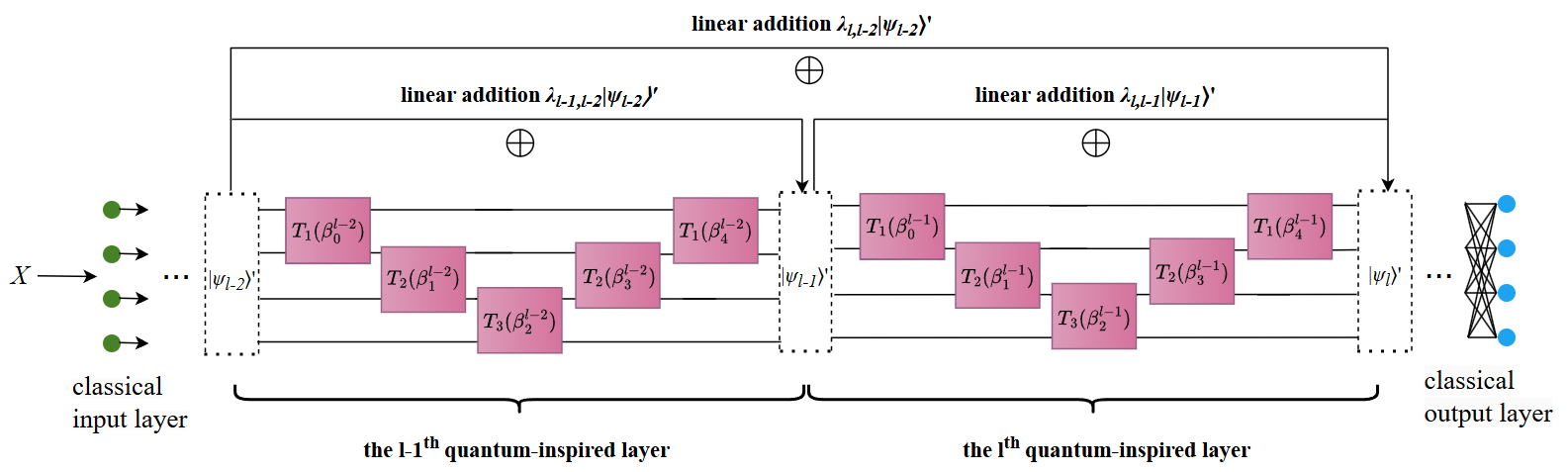}%
		\label{QDFNN}
	}
	\caption{Frameworks of QRFNN and QDFNN. (a) QRFNN. (b) QDFNN.}
\end{figure*}
\section{Hybrid Quantum-inspired Resnet and Densenet}
We describe our hybrid models with specific architectures in this section, which include layer details of QRFNN and QDFNN and parameter learning of QRFNN and QDFNN.
\subsection{Layer details of QRFNN and QDFNN}
Fig.3 demonstrates the QRFNN and QDFNN structures of 2 qubits with 2 layers, where each line represents one-dimensional vector space of a quantum state. The green, blue circles denote the input and output neurons respectively. To contemplate the general occasions, suppose we build $L$ quantum-inspired layers, the quantum-inspired part, as hidden layers, in QRFNN and QDFNN rather than MLP hidden layers with neurons. So there are $L+2$ layers totally in both QRFNN and QDFNN. We set the quantum states in the quantum-inspired layers are $N$-dimensional with $n$ qubits utilized for QRFNN and QDFNN $(2^{n-1} \leq N \leq 2^n)$. And $T_{i}(\theta^l_r) \in \mathbb{R}^{N \times N}$. We also suppose the dimension of the input, $X =(x_0,x_1,...,x_{N-1})^\top$, is $N$. In terms of $r$ and $i$ in Eq.(3) in this section, we also have:
\begin{equation}
	r=
	\begin{cases}
		i-1& \text{ $ 0 \leq r \leq N-2 $ } \\
		2N-3-i& \text{ $ N-2 < r \leq 2N-4. $ }
	\end{cases}
\end{equation}

Therefore, the row and parameter indices in Eq.(3) can be represented with only one variable simultaneously. For multiplication of the $T$ matrices in the $l^{th}$ quantum-inspired layer, we define:
\begin{subequations}
	\begin{align}
		\label{deqn_ex1a}
		\prod \limits_{\substack{j=0}}^{\substack{j=N}}T_{j+1}(\alpha_{j}^{l-1}) \coloneqq T_{1}(\alpha_{0}^{l-1})T_{2}(\alpha_{1}^{l-1})...T_{N+1}(\alpha_{N}^{l-1}), \\
		\prod \limits_{\substack{j=0}}^{\substack{j=N}}T_{j+1}(\beta_{j}^{l-1}) \coloneqq T_{1}(\beta_{0}^{l-1})T_{2}(\beta_{1}^{l-1})...T_{N+1}(\beta_{N}^{l-1}).
	\end{align}
\end{subequations}

For the $l^{th}$ quantum-inspired layer, we define:
\begin{subequations}
	\begin{small}
	\begin{align}
		U_{l-1} \coloneqq (\prod \limits_{\substack{j=0}}^{\substack{j=N-2}}T_{j+1}(\alpha_{j}^{l-1})) \times (\prod \limits_{\substack{j=N-1}}^{\substack{j=2N-4}}T_{2N-3-j}(\alpha_{j}^{l-1})),\\
		U_{l-1}^{\prime} \coloneqq (\prod \limits_{\substack{j=0}}^{\substack{j=N-2}}T_{j+1}(\beta_{j}^{l-1})) \times (\prod \limits_{\substack{j=N-1}}^{\substack{j=2N-4}}T_{2N-3-j}(\beta_{j}^{l-1})),
	\end{align}
    \end{small}
\end{subequations}
where ${\alpha}^{l-1}_j$ and ${\beta}^{l-1}_j$ are the $(j+1)^{th}$ parameters in the $l^{th}$ layer of QRFNN and QDFNN separately. And $U_{l-1}$, $U_{l-1}^{\prime}$ are the multiplications of all the unitary gates in the $l^{th}$ quantum-inspired layer of QRFNN and QDFNN separately. And the general framework of the $l^{th}$ quantum-inspired layer is shown as Fig.4.\\
\begin{figure}[!t]
	\centering
	\includegraphics[width=3.5in]{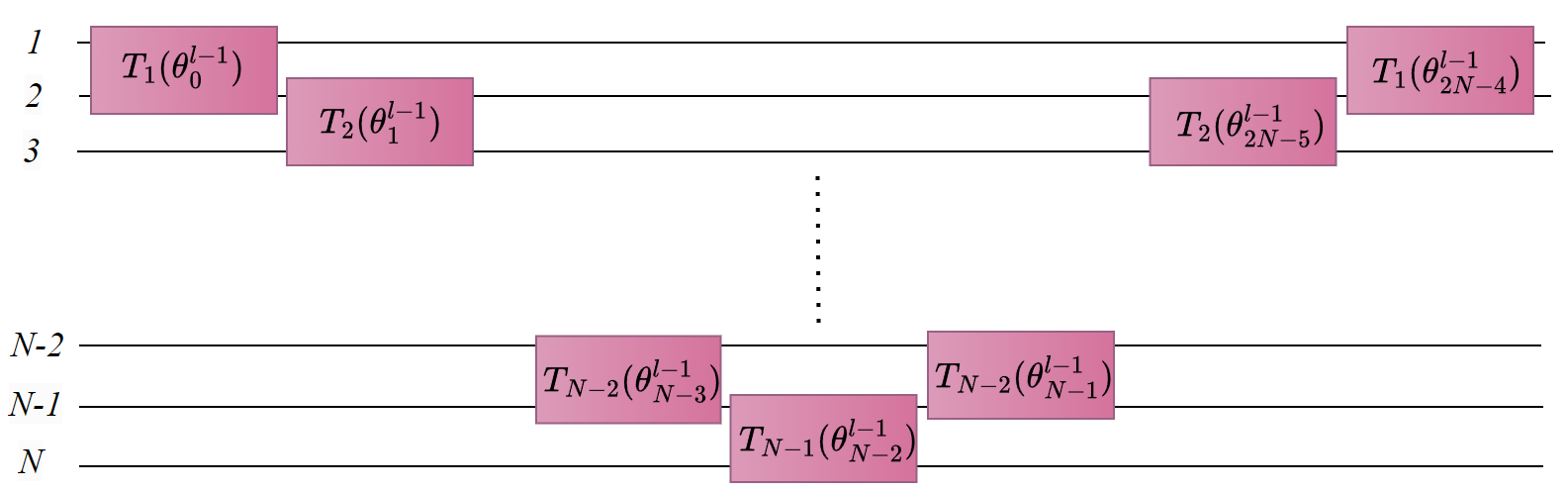}
	\caption{General structure of the $l^{th}$ quantum-inspired layer. Here the parameter is represented as $\theta$, but in QRFNN and QDFNN, we use $\alpha$ and $\beta$ as parameters respectively. And each line represents one-dimensional vector space of the states. Furthermore, the "V" shape helps feature learning of each category of the dataset. On one hand, if the circuit shape is not the symmetrical "V" shape, there could be imbalance of the number of parameters of each category of a dataset for feature learning. On the other hand, each unitary gate is a sparse matrix with only one parameter, which ensure the convenience of computation and guarantee that over-parameterization does not occur.}
\end{figure}

As for the evolution process, firstly, $X$ is input into the neurons of the input layer, then we encode the information of $X$ into the amplitude of $\ket{\psi_0}$ of QRFNN and $\ket{\psi_0}^{\prime}$ of QDFNN:
\begin{subequations}
	\begin{align}
	\label{deqn_ex1a}
	\eta &= \sqrt{\sum_{k=0}^{N-1}x_k^2},\\
	\ket{\psi_0} &= \sum_{k=0}^{N-1}\frac{x_k}{\eta}\ket{k},\\
	\ket{\psi_0}^{\prime} &= \sum_{k=0}^{N-1}\frac{x_k}{\eta}\ket{k^{\prime}},
    \end{align}
\end{subequations}
where we represent $\ket{k}$ and $\ket{k^{\prime}}$ using the binary representation $k = k_{n-1}k_{n-2}...k_0$. More formally, $k = \sum_{j=0}^{n-1}k_j 2^{j}$. \\
Secondly, as shown in Fig.3, in terms of QRFNN:
\begin{equation}
	\ket{\psi_l} = U_{l-1}\ket{\psi_{l-1}} \oplus \lambda_{l-1}\ket{\psi_{l-1}}\quad(1 \leq l \leq L).
\end{equation}

While for QDFNN:
\begin{equation}
	\ket{\psi_l}^{\prime}= U_{l-1}^{\prime}\ket{\psi_{l-1}}^{\prime} \oplus \sum^{j=l-1}_{j=0}(\lambda_{l,j}\ket{\psi_{j}}^{\prime})\quad(1 \leq l \leq L),
\end{equation}
 and $\ket{\psi_l}, \ket{\psi_l}^{\prime}$ are respectively the output states of the $l^{th}$ quantum-inspired layers of QRFNN and QDFNN. $\oplus$ denotes the element-wise addition mechanism. The terms $U_{l-1}\ket{\psi_{l-1}}$ and $U_{l-1}^{\prime}\ket{\psi_{l-1}}^{\prime}$ indicate the quantum-inspired parts of the two models obey laws of circuit models \cite{22}. However, because our hybrid models also follow the deep learning laws, $\ket{\psi_l}$ and $ \ket{\psi_l}^{\prime}$ can be also respectively represented as column vectors $O_l$ and $O_l^{\prime}$ (see supplementary materials section 2). For QRFNN, $O_l = (o_{l,0},o_{l,1},...,o_{l,N-1})^\top$. For QDFNN, $ O_l^{\prime} = (o_{l,0}^{\prime},o_{l,1}^{\prime},...,o_{l,N-1}^{\prime})^\top$. And $o_{l,j},o_{l,j}^{\prime}$ denote the output at the $(j+1)^{th}$ dimensional index of the $l^{th}$ quantum-inspired hidden layer of the two models respectively. And we have:
 \begin{subequations}
 	\begin{align}
 		o_{l,j}& = \sum^{N-1}_{j=0}(v_j\cdot o_{l-1,j}),\\
 		o_{l,j}^{\prime}& = \sum^{N-1}_{j=0}(v_j^{\prime}\cdot o_{l-1,j}^{\prime}),
 	\end{align}
 \end{subequations}
where $v_j$, the coefficient of $o_{l-1,j}$, is the linear summations of many terms, where each term is the multiplication of $\lambda_{l-1}, \sin(\alpha^{l-1}_r), \cos(\alpha^{l-1}_r)$ and the input data. Similarly, $v_j^{\prime}$, the coefficient of $o_{l-1,j}^{\prime}$, is also the linear summations of many terms, where each term is the multiplication of $\lambda_{l,l-1}, \sin(\beta^{l-1}_r), \cos(\beta^{l-1}_r)$ and the input data. The output of the last quantum-inspired layer of QRFNN and QDFNN is $O_L$ and $O_L^{\prime}$ respectively. Eventually, the final outputs will be:
\begin{subequations}
	\begin{align}
		&O_{L+1} = \sigma((W_{L+1}\cdot O_L)^\top+ B_{L+1}),\\
		&O_{L+1}^{\prime} = \sigma((W_{L+1}^{\prime}\cdot O_L^{\prime})^\top+ B_{L+1}^{\prime}),
	\end{align}
\end{subequations}
where $\sigma$ denotes any activation function, $W_{L+1}$ and $W_{L+1}^{\prime}$ mean the weight matrices of the output layers of QRFNN and QDFNN separately, while $B_{L+1},B_{L+1}^{\prime}$ represent biases of QRFNN and QDFNN respectively. Therefore, our hybrid neural networks are constituted by quantum-inspired hidden layers, as well as pure classical input and output layers.

\textcolor{black}{Besides that, MNIST, FASHIONMNIST and CIFAR image datasets are utilized for validation of QRCNN and QDCNN. Compared with QRFNN and QDFNN, they have one more convolutional module, a few convolutional layers with parallel spatial and channel attentions respectively. Hence, high dimensional features of the image datasets are firstly extracted by parameters in convolutional kernels and pooling operation, which reduces the feature dimension of the data. Then the tensor data is flattened to vectors before they are transferred to quantum-inspired layers. The vectors are encoded into the amplitudes of the quantum states. After the feature learning of quantum-inspired layers, the vectors are transmitted to the output layers. Section 6 of the supplementary materials provides detailed frameworks of QRCNN and QDCNN. The discrepancy between the CNNs and QRCNN only lies in the fully-connected layers in CNNs and the quantum-inspired layers in QRCNN.} 
\subsection{Parameter learning of QRFNN and QDFNN}
In the section 1 of the supplementary material, the framework of the MLP is also given. By comparison, it is found that the similarity of QRFNN, QDFNN and MLP lies in the feedforward layer-by-layer transfer of features. What's more, the classical output layer with activation functions and biases further enhances the ability of our hybrid models for nonlinear problems \cite{2}. It also makes sense to activate the quantum-inspired layers with assorted activation functions. The advantage of the classical input and output layers of QRFNN and QDFNN is the convenience of changing the number of neurons. Hence, our hybrid models are capable of allowing inputs and outputs of different dimensions. In addition, on the basis of the chain rule \cite{2}, the gradients of loss functions to the parameter $\alpha_r^{l-1}$ and $\beta_r^{l-1}$ are given:
\begin{subequations}
	\begin{align}
		\frac{\partial (loss)}{\partial \alpha_r^{l-1}}& = \frac{\partial (loss)}{\partial O_{L+1}}\cdot (\prod \limits_{j=0}^{L-l} \frac{\partial O_{L+1-j}}{\partial O_{L-j}}) \cdot \frac{\partial O_l}{\partial \alpha_r^{l-1}},\\
		\frac{\partial (loss^{\prime})}{\partial \beta_r^{l-1}}& = \frac{\partial (loss^{\prime})}{\partial O_{L+1}^{\prime}}\cdot (\prod \limits_{j=0}^{L-l} \frac{\partial O_{L+1-j}^{\prime}}{\partial O_{L-j}^{\prime}}) \cdot \frac{\partial O_l^{\prime}}{\partial \beta_r^{l-1}},
	\end{align}
\end{subequations}
where $loss$ and $loss^{\prime}$ denote the loss functions of QRFNN and QDFNN separately. The terms, $\frac{\partial O_{L+1-j}}{\partial O_{L-j}}$ and $\frac{\partial O_{L+1-j}^{\prime}}{\partial O_{L-j}^{\prime}}$ $(0 \leq j \leq L-l$), are derivative matrices. And each element in each of the matrices involves additions and multiplications of many sine and cosine functions. Because the ranges of sine and cosine functions are both [-1,1], there are limitations of the absolute value of each element in each of the matrices, which makes the absolute value of each element not very large. As a result, the absolute values of $\frac{\partial (loss)}{\partial \alpha_r^{l-1}}$ and $\frac{\partial (loss)}{\partial \beta_r^{l-1}}$ will be not very large, either. On the other hand, a common problem at deep learning region, gradient explosion, is caused by very large values of the gradients of the loss functions to the parameters \cite{2}. As a result, the parameters in the neural networks are unable to be updated and the neural networks are not capable of learning features of new data. However, our hybrid models may be able to prevent gradient explosion through this theoretical analysis. More details of the illustration of our hybrid models to prevent gradient explosion are in the section 4 of the supplementary material.
\section{Numerical experiments}
In this section, we validate the generalization power of our hybrid models with various datasets. We also illustrate the robustness of our hybrid models with various parameter attacks utilizing noise. The programming architecture we use is Pytorch. \textcolor{black}{All the experiments are from single runs.} Two MLPs with different activation functions are used for comparison with QRFNN and QDFNN. The details of the MLPs are in section 1 and 10 of the supplementary material. HQNet and Two CNNs with different activation functions are used for comparison with QRCNN and QDCNN on generalization power and robustness test. \textcolor{black}{Moreover, In terms of the original HQNet proposed in 2024, we contemplate it as the state-of-the-art hybrid models for comparison since its hybrid framework is similar with those of our QRCNN and QDCNN \cite{2,49}. Another reason lies in its more outstanding performance on several medical image classification tasks than numerous state-of-the-art classical convolutional networks such as DenseNet-121, MBTFCN, AG-CNN, etc. For instance, the experimental results on classification of BT-large-4c dataset indicate that while the accuracy of the advanced EfficientNetB0 and MobileNet is respectively 96.48\% and 95.10\%, that of the original HQNet is 97.55\%. Simultaneously, the network model parameters and memory of the original HQNet is much less than that of EfficientNetB0 and MobileNet \cite{49}. When it comes to the detailed architecture of the original HQNet, it is comprised of a hybrid residual module, a feature extraction and recombination module, a random quantum circuit module, and a classification module. And as for the HQNet in our experiments, we maintain the four modules but reduce the size of the random circuit model of the original HQNet from 4 qubits to 2 qubits to apply to smaller-size image classification. We use ${\rm R}_{\rm x}$ and ${\rm CR}_{\rm x}$ gate to construct the random circuit in our HQNet (see section 13 of the supplementary material) \cite{49}. We also alter the number of the output neuron from 2 to 4 in the classification module for our multi-classification problem. What's more, ref. \cite{8} and \cite{23} have shown neural networks with different residual mappings, all of which demonstrate outstanding performances. Hence, to further verify the ability of the residual and dense mapping our proposed hybrid models, we devise ResQNet1 and ResQNet2 by adjusting the residual mapping of QRCNN for comparison. The frameworks of ResQNet1, ResQNet2 and QRCNN are almost identical with the only discrepancy being the residual mapping in the quantum-inspired part. While one quantum-inspired layer constitute the residual mapping of QRCNN, that of ResQNet1 and ResQNet2 is comprised of all quantum-inspired layers and two quantum-inspired layers respectively.} In addition, when it comes to the two classical CNNs, they consist of convolutional layers and fully connected layers with MLPs. And QRCNN is comprised of residually-connected convolutional layers and fully connected layers with QRFNN, and QDCNN is comprised of densely-connected convolutional layers and fully connected layers with QDFNN. The convolutional parts in CNNs and QRCNN are identical. And we set most of the hyperparameters including the number of fully connected layers of our HQNet, CNNs, QDCNN and QRCNN, learning rate, etc., the same to compare these models more accurately (see section 10 of the supplementary material for hyperparameter setting). Moreover, an ablation study is also conducted to test the superiority of the residual and dense connections. The traditional quantum-inspired neural network without residual or dense connection is utilized for comparison in it. All the results in this chapter are testing results. 
\subsection{Generalization power test}
\begin{table*}[!t]
	\centering
	\renewcommand{\arraystretch}{0.75}
	\caption{Results on FASHIONMNIST with gaussian symmetrical noise}
	\label{table_example}
	\centering
	\scalebox{0.8}{
	\begin{tabular}{c c c c c c}
		\toprule
		  & QRCNN & CNN(leaky-relu) & QDCNN & CNN(rot-relu) \\
		\midrule
		P-R curve area & 0.9730 & 0.9766 & 0.9768 & \textbf{0.9788} \\
		ROC curve area & 0.9892 & 0.9911 & 0.9910 & \textbf{0.9917} \\
		Accuracy       & 0.9284$\pm$0.0179 & 0.9299$\pm$0.0177 & \textbf{0.9316$\pm$0.0175} & 0.9310$\pm$0.0176 \\
		\bottomrule
	\end{tabular}}
\end{table*}

\begin{table*}[!t]
	\renewcommand{\arraystretch}{0.75}
	\centering
	\caption{Results of category 0 of FASHIONMNIST with gaussian symmetrical noise}
	\label{table_example}
	\centering
	\scalebox{0.8}{
	\begin{tabular}{c c c c c c}
		\toprule
		&   & QRCNN & CNN(leaky-relu) & QDCNN & CNN(rot-relu) \\
		%\hline
		%\cline{3-5} 
		\hline
		\multicolumn{1}{c|}{\multirow{3}{*}{variance}} & recall & 1.290E-02 & 1.163E-02 & \textbf{9.408E-03} & 1.437E-02 \\ 
		%\hline
		\multicolumn{1}{c|}{} & precision & 1.110E-02 & 1.062E-02 & \textbf{1.001E-02} & 1.008E-02 \\
		%\cline{3-5}
		\multicolumn{1}{c|}{}   & f1 score  & 6.558E-03 & 6.547E-03 & \textbf{5.022E-03} & 7.114E-03  \\ 
		\hline
		\multicolumn{1}{c|}{\multirow{3}{*}{mean value}} & recall & 0.9220 & 0.9206 & \textbf{0.9331} & 0.9144 \\ 
		%\cline{3-5}
		%\hline
		\multicolumn{1}{c|}{} & precision & 0.8982 & 0.9102 & 0.9076 & \textbf{0.9108} \\
		%\hline
		\multicolumn{1}{c|}{}   & f1 score  & 0.9018  & 0.9091 & \textbf{0.9136} & 0.9049    \\ 
		\bottomrule
	\end{tabular}}
\end{table*}

\begin{table*}[!t]
	\renewcommand{\arraystretch}{0.75}
	\centering
	\caption{Results of category 1 of FASHIONMNIST with gaussian symmetrical noise}
	\label{table_example}
	\centering
	\scalebox{0.8}{
	\begin{tabular}{c c c c c c}
		\toprule
		&   & QRCNN & CNN(leaky-relu) & QDCNN & CNN(rot-relu) \\
		%\hline
		%\cline{3-5} 
		\hline
		\multicolumn{1}{c|}{\multirow{3}{*}{variance}} & recall & 4.551E-03 & 6.904E-03 & \textbf{3.570E-03} & 5.179E-03 \\ 
		%\hline
		\multicolumn{1}{c|}{} & precision & 3.872E-03 & \textbf{3.742E-03} & 3.936E-03 & 3.862E-03 \\
		%\cline{3-5}
		\multicolumn{1}{c|}{}   & f1 score  & 2.385E-03 & 3.470E-03 & \textbf{2.070E-03} & 3.168E-03  \\ 
		\hline
		\multicolumn{1}{c|}{\multirow{3}{*}{mean value}} & recall & 0.9751 & 0.9702 & 0.9801 & \textbf{0.9805} \\ 
		%\cline{3-5}
		%\hline
		\multicolumn{1}{c|}{} & precision & 0.9642 & 0.9647 & 0.9655 & \textbf{0.9657} \\
		%\hline
		\multicolumn{1}{c|}{}   & f1 score  & 0.9672  & 0.9638 & \textbf{0.9705} & 0.9703  \\ 
		\bottomrule
	\end{tabular}}
\end{table*}

\begin{table*}[!t]
	\renewcommand{\arraystretch}{0.75}
	\centering
	\caption{Results of category 2 of FASHIONMNIST with gaussian symmetrical noise}
	\label{table_example}
	\centering
	\scalebox{0.8}{
	\begin{tabular}{c c c c c c}
		\toprule
		&   & QRCNN & CNN(leaky-relu) & QDCNN & CNN(rot-relu) \\
		%\hline
		%\cline{3-5} 
		\hline
		\multicolumn{1}{c|}{\multirow{3}{*}{variance}} & recall & 1.210E-02 & 1.408E-02 & \textbf{9.901E-03} & 1.254E-02 \\ 
		%\hline
		\multicolumn{1}{c|}{} & precision & 5.680E-03 & 6.067E-03 & \textbf{5.179E-03} & 7.347E-03 \\
		%\cline{3-5}
		\multicolumn{1}{c|}{}   & f1 score  & 5.349E-03 & 6.463E-03 & \textbf{4.800E-03} & 5.819E-03  \\ 
		\hline
		\multicolumn{1}{c|}{\multirow{3}{*}{mean value}} & recall & 0.9140 & 0.9112 & \textbf{0.9157} & 0.9125 \\ 
		%\cline{3-5}
		%\hline
		\multicolumn{1}{c|}{} & precision & \textbf{0.9599} & 0.9472 & 0.9590 & 0.9426 \\
		%\hline
		\multicolumn{1}{c|}{}  & f1 score  & 0.9311  & 0.9227 & \textbf{0.9328} & 0.9213    \\ 
		\bottomrule
	\end{tabular}}
\end{table*}

\begin{table*}[!t]
	\renewcommand{\arraystretch}{0.75}
	\centering
	\caption{Results of category 3 of FASHIONMNIST with gaussian symmetrical noise}
	\label{table_example}
	\centering
	\scalebox{0.8}{
	\begin{tabular}{c c c c c c}
		\toprule
		&   & QRCNN & CNN(leaky-relu) & QDCNN & CNN(rot-relu) \\
		%\hline
		%\cline{3-5} 
		\hline
		\multicolumn{1}{c|}{\multirow{3}{*}{variance}} & recall & 1.628E-02 & \textbf{1.320E-02} & 1.363E-02 & 1.556E-02 \\ 
		%\hline
		\multicolumn{1}{c|}{} & precision & 7.740E-03 & 8.798E-03 & \textbf{6.703E-03} & 9.586E-03 \\
		%\cline{3-5}
		\multicolumn{1}{c|}{}   & f1 score  & 7.353E-03 & 6.443E-03 & \textbf{6.118E-03} & 7.401E-03  \\ 
		\hline
		\multicolumn{1}{c|}{\multirow{3}{*}{mean value}} & recall & 0.8960 & 0.9023 & \textbf{0.9047} & 0.8930 \\ 
		%\cline{3-5}
		%\hline
		\multicolumn{1}{c|}{} & precision & 0.9312 & 0.9254 & \textbf{0.9391} & 0.9234 \\
		%\hline
		\multicolumn{1}{c|}{}   & f1 score  & 0.9056  & 0.9068 & \textbf{0.9154} & 0.9003    \\ 
		\bottomrule
	\end{tabular}}
\end{table*}

\begin{table}[!t]
	\centering
	\renewcommand{\arraystretch}{0.75}
	\caption{Accuracy on FASHIONMNIST with three symmetrical noises}
	\label{table_example}
	\centering
	\scalebox{0.8}{
	\begin{tabular}{c c c c}
		\toprule
		 & QRCNN & HQNet & QDCNN \\
		\midrule
		  mixed noise & 0.9003$\pm$0.0105 & 0.9011$\pm$0.0102 & \textbf{0.9280$\pm$0.0094} \\
		  gaussian noise & 0.9157$\pm$0.0101 & 0.9194$\pm$0.0100 & \textbf{0.9294$\pm$0.0096} \\
		  uniform noise & 0.9165$\pm$0.0103 & 0.9096$\pm$0.0106 & \textbf{0.9380$\pm$0.0093} \\
		\bottomrule
	\end{tabular}}
\end{table}

\begin{table}[!t]
	\centering
	\renewcommand{\arraystretch}{0.75}
	\caption{Accuracy on MNIST with three symmetrical noises}
	\label{table_example}
	\centering
	\scalebox{0.8}{
		\begin{tabular}{c c c c}
			\toprule
			& QRCNN & HQNet & QDCNN \\
			\midrule
			mixed noise & 0.9472$\pm$0.0105 & 0.9532$\pm$0.0095 & \textbf{0.9680$\pm$0.0088} \\
			gaussian noise & 0.9506$\pm$0.0095 & 0.9491$\pm$0.0097 & \textbf{0.9794$\pm$0.0082} \\
			uniform noise & 0.9412$\pm$0.0106 & 0.9405$\pm$0.0107 & \textbf{0.9711$\pm$0.0083} \\
			\bottomrule
	\end{tabular}}
\end{table}

\begin{table}[!t]
	\centering
	\renewcommand{\arraystretch}{0.75}
	\caption{Accuracy on FASHIONMNIST with three asymmetrical noises}
	\label{table_example}
	\centering
	\scalebox{0.8}{
	\begin{tabular}{c c c c}
		\toprule
		& QRCNN & HQNet & QDCNN \\
		\midrule
		mixed noise & 0.8957$\pm$0.0103 & 0.8787$\pm$0.0111 & \textbf{0.9270$\pm$0.0096} \\
		gaussian noise & 0.9003$\pm$0.0105 & 0.9096$\pm$0.0107 & \textbf{0.9201$\pm$0.0097} \\
		uniform noise & 0.9241$\pm$0.0092 & 0.9190$\pm$0.0096 & \textbf{0.9422$\pm$0.0086} \\
		\bottomrule
	\end{tabular}}
\end{table}

\begin{table}[!t]
	\centering
	\renewcommand{\arraystretch}{0.75}
	\caption{Accuracy on MNIST with three asymmetrical noises}
	\label{table_example}
	\centering
	\scalebox{0.8}{
		\begin{tabular}{c c c c}
			\toprule
			& QRCNN & HQNet & QDCNN \\
			\midrule
			mixed noise & 0.9494$\pm$0.0098 & 0.9305$\pm$0.0101 & \textbf{0.9682$\pm$0.0086} \\
			gaussian noise & 0.9311$\pm$0.0100 & 0.9435$\pm$0.0095 & \textbf{0.9603$\pm$0.0087} \\
			uniform noise & 0.9487$\pm$0.0098 & 0.9475$\pm$0.0094 & \textbf{0.9622$\pm$0.0089} \\
			\bottomrule
	\end{tabular}}
\end{table}
Generalization power is appraised by accuracy, precision, recall, f1 score, P-R curve area and ROC curve area \cite{2}. Precision refers to the proportion of samples that are actually positive among all samples predicted to be positive by the model, while recall represents the proportion of samples that the model can correctly predict to be positive among all samples that are actually positive \cite{1}. And f1 score is an important indicator for balancing recall and precision. Their definitions are follows:
\begin{equation}
	\label{deqn_ex1a}
	Accuracy = \frac{TP+TN}{TP+TN+FP+FN}.
\end{equation}
\begin{equation}
	\label{deqn_ex1a}
	Recall = \frac{TP}{TP+FN}.
\end{equation}
\begin{equation}
	\label{deqn_ex1a}
	Precision = \frac{TP}{FP+TP}.
\end{equation}
\begin{equation}
	\label{deqn_ex1a}
	f1\; score = \frac{2\times Precision \times Recall}{Precision + Recall},
\end{equation}
where True Positive $(TP)$ denotes the number of samples of a category truly predicted as positive if we regard this category as positive in a dataset and other categories in the dataset negative. True Negative $(TN)$ is the number of samples predicted as negative that is negative. False Positive $(FP)$ is the number of samples that are incorrectly predicted to be negative but are positive in fact. False Negative $(FN)$ represents the number of samples that are falsely predicted to be positive but are actually negative. The P-R curve plots precision against recall at different classification thresholds, while the ROC curve plots true positive rate against false positive rate at assorted classification thresholds. They are expected to be close to 1 \cite{2}. Furthermore, we utilize iris data with three categories and MNIST, FASHIONMNIST, CIFAR100 image datasets with each image dataset comprised of four categories. The iris data is 4-dimensional, the size of MNIST and FASHIONMNIST is 28 $\times$ 28 $\times$ 1, while that of CIFAR dataset is 32 $\times$ 32 $\times$ 3. In terms of each dataset, we set 75\% of it as training set and 25\% of it as testing set. And the proportion of each category of data in each dataset is approximately the same. To emulate real-world scenarios more closely, the noises added are pure symmetrical and pure asymmetrical gaussian noises, pure symmetrical and pure asymmetrical uniform noises, and mixed noises with symmetrical and asymmetrical distributions, a total of six cases \cite{41,42,43}. The symmetrical noise means the noise data distribution which is symmetrical with respect to the y-axis in a Cartesian coordinate system (see supplementary material section 12) \cite{42}. These noises used in the paper are significant in engineering control \cite{41}. There are more types of asymmetrical noise than symmetrical noise \cite{41}. Therefore, in the engineering field, the probability of asymmetrical noise occurring is more than that of symmetrical noise \cite{41,44}. Moreover, to add more uncertainties to simulate the situations of real world, the amount and the position of the noise added to the data are both random in each group of the experiment \cite{42}. And in each group of generalization power experiment, the noisy dataset utilized for training and testing all the models is exactly identical. For instance, the four models are evaluated with identical FASHIONMNIST dataset with gaussian symmetrical noise in table 1.
\subsubsection{Test on datasets with symmetrical noises}
%\begin{figure}[!t]
%	\centering
%	\includegraphics[width=3.5in]{fashion_gaussian_symm.png}
%	\caption{Results with gaussian symmetrical noise in FASHIONMNIST dataset. We utilize two pure classical CNNs for comparison with the hybrid models, one of which is activated by leaky-relu function. And the activation function in the other pure classical CNN is rot-relu (see supplementary material part 3).}
%\end{figure}
Table 1 to Table 5 demonstrate the outcomes of FASHIONMNIST with pure gaussian symmetrical noise (more results of other datasets in supplementary materials section 7). The results include the P-R and ROC curve area and accuracy. They also incorporate variances and mean values of precision, recall and f1 score of the four categories, from category 0 to category 3. And the category Y (Y=0,1,2,3) means a category labeled as the number Y. The mean value is the average values of precision, recall and f1 score of all the samples in a certain category from the $100^{th}$ epoch to the $300^{th}$ epoch after convergence, which implies the learning ability of models for samples of different categories. The variance represents that of precision, recall and f1 score of all the samples in a certain category from the $100^{th}$ epoch to the $300^{th}$ epoch after convergence, which indicates the degree of fluctuation of the indicators, reflecting the stability of the models. As for Table 1, while the accuracy of QRCNN and the two CNNs is 92.84\%$\pm$0.96\% , 92.99\%$\pm$0.94\%  and 93.10\%$\pm$0.93\% , that of QDCNN is 93.16\%$\pm$0.93\% , which is slightly higher than that of QRCNN and CNNs. However, the accuracy difference between these models is exceedingly small. And the differences of P-R and ROC curve area between the four models are very small as well. Table 2 shows the variances and mean values of recall, precision and f1 score of category 0 of noisy FASHIONMNIST with category 0 considered as positive and other categories regarded as negative. Table 3 shows the variances and mean values of recall, precision and f1 score of category 1 of noisy FASHIONMNIST with category 1 considered as positive and other categories regarded as negative. Table 4 shows the variances and mean values of recall, precision and f1 score of category 2 of noisy FASHIONMNIST with category 2 considered as positive. Table 5 shows the variances and mean values of recall, precision and f1 score of category 3 of noisy FASHIONMNIST with category 3 considered as positive. As for the Table 2 to Table 5, there are only very small discrepancy of the variances and mean values of the three metrics between the four models. And the outcomes in section 7 in the supplementary material also indicate the exceedingly small discrepancy of those. Therefore, our hybrid models could not outperform pure classical models in all these indicators. They only perform approximately on par with classical CNNs on datasets with symmetrical noises on various metrics. What's more, Table 6 and 7 demonstrate accuracy of HQNet, QRCNN as well as QDCNN on noisy FASHIONMNIST and MNIST respectively. The results point the slight superiority of QDCNN over QRCNN and HQNet. While QRCNN shows approximately as high accuracy as the HQNet model, more significantly, the accuracy of QDCNN is about 2\%-3\% higher than HQNet and QRCNN models, which shows the ascendancy of the dense connection of the convolutional and quantum-inspired layers in QDCNN.
\subsubsection{Test on datasets with asymmetrical noises}
\begin{table*}[!t]
	\centering
	\renewcommand{\arraystretch}{0.75}
	\caption{Accuracy on CIFAR10 with three asymmetrical noises}
	\label{table_example}
	\centering
	\scalebox{0.8}{
		\begin{tabular}{c c c c c c}
			\toprule
			noise type& ResQNet1&ResQNet2 & QRCNN & QDCNN   \\
			\midrule
			uniform noise& 78.04\%$\pm$1.29\%&77.49\%$\pm$1.47\% & 76.82\%$\pm$1.62\% & \textbf{79.33\%$\pm$1.16\%}  \\
			gaussian noise&75.98\%$\pm$1.71\%&76.08\%$\pm$1.67\% & 76.17\%$\pm$1.66\% & \textbf{77.13\%$\pm$1.64\%}   \\
			mixed noise& 79.60\%$\pm$1.23\%&78.24\%$\pm$1.32\% & \textbf{80.01\%$\pm$1.21\%} & 79.96\%$\pm$1.22\%  \\
			\bottomrule
	\end{tabular}}
\end{table*}

\begin{table*}[!t]
	\centering
	\renewcommand{\arraystretch}{0.75}
	\caption{Accuracy on MNIST with three asymmetrical noises}
	\label{table_example}
	\centering
	\scalebox{0.8}{
		\begin{tabular}{c c c c c c}
			\toprule
			noise type& ResQNet1&ResQNet2 & QRCNN & QDCNN  \\
			\midrule
			uniform noise&  94.66\%$\pm$0.70\%&\textbf{96.94\%$\pm$0.58\%} & 95.95\%$\pm$0.62\% & 96.71\%$\pm$0.59\%  \\
			gaussian noise& 93.86\%$\pm$0.78\%&95.42\%$\pm$0.66\% & 95.05\%$\pm$0.69\% & \textbf{96.22\%$\pm$0.59\%}  \\
			mixed noise& 93.23\%$\pm$0.85\%&93.97\%$\pm$0.79\% & 94.01\%$\pm$0.76\% & \textbf{94.73\%$\pm$0.77\%}  \\
			\bottomrule
	\end{tabular}}
\end{table*}
Table 8 and 9 show the accuracy of HQNet, QRCNN and QDCNN on noisy FASHIONMNIST and MNIST respectively. The outcomes also reveal slightly more remarkable performance of QDCNN with 2\%-3\% higher accuracy over HQNet and QRCNN. Other results of the comparison between our hybrid models and classical MLPs and CNNs are shown in the section 7 of the supplementary material. On the whole, the outcomes indicate that our hybrid models possess roughly the same level of generalization capability as MLPs or CNNs in recognizing data containing asymmetrical noises.

\textcolor{black}{What's more, Table 10 and 11 showcase the accuracy of ResQNet1, ResQNet2 and our hybrid models on CIFAR10 and MNIST datasets with the three asymmetrical noises. The outcomes reveal a little bit more prominent performance of our hybrid models over ResQNet1 and ResQNet2, which demonstrates the slightly more prominent function of the residual mapping with one quantum-inspired layer.}

\subsection{Robustness test}
Robustness is appraised by the same metrics in generalization power part. And in this part, we use the same datasets as in generalization power part. For QDFNN, QRFNN and the MLPs, in each group of robustness experiments, we randomly choose one layer and attack some parameters randomly in the layer utilizing the six noises in generalization power test. The layer indices and the number of parameters that are chosen for attacking in the layer of the three models are identical. In terms of QRCNN, QDCNN, CNNs, HQNet, ResQNet1 and ResQNet2, since we hope to evaluate the robustness of the fully-connected layers of these models, we randomly choose a fully-connected layers in the CNNs, a quantum-inspired layer in QRCNN, QDCNN, ResQNet1 and ResQNet2, a quantum layer in HQNet for attacking. And we ensure the layer indices and the number of parameters attacked in the layer in CNNs is also exactly the same as that in HQNet, QDCNN, QRCNN, ResQNet1 and ResQNet2. In each group of robustness experiment, the noise used to attack different models is identical. \textcolor{black}{For example, in terms of each of the three models in Fig.(5a), one random quantum or one quantum-inspired layer with the same index is attacked by identical symmetrical mixed noise in form \ding{172}.} Nonetheless, the noise used for attacking between different groups of experiments and the number of parameters chosen in one layer for attacking between various groups of experiments are different. There are two common attacking forms: \ding{172}  $\sigma(\theta+\epsilon)$ and \ding{173} $\sigma(\theta)+\epsilon$, where $\theta$ is the parameter and $\epsilon$ is the noise. $\sigma$ represents activation operations in pure classical models or sine and cosine functions in our hybrid models or HQNet\cite{41,45,46,47}. The test of robustness is divided into symmetrical noise attack with form \ding{172},\ding{173} and asymmetrical noise attack with form \ding{172},\ding{173}. The accuracy value is the average one after the models reach convergence. The link to the program codes is on the last page of this paper.
\subsubsection{Test under symmetrical noise attacks}
\begin{figure*}[!t]
	\centering
	\subfloat[\centering]{\includegraphics[width=3.6in]{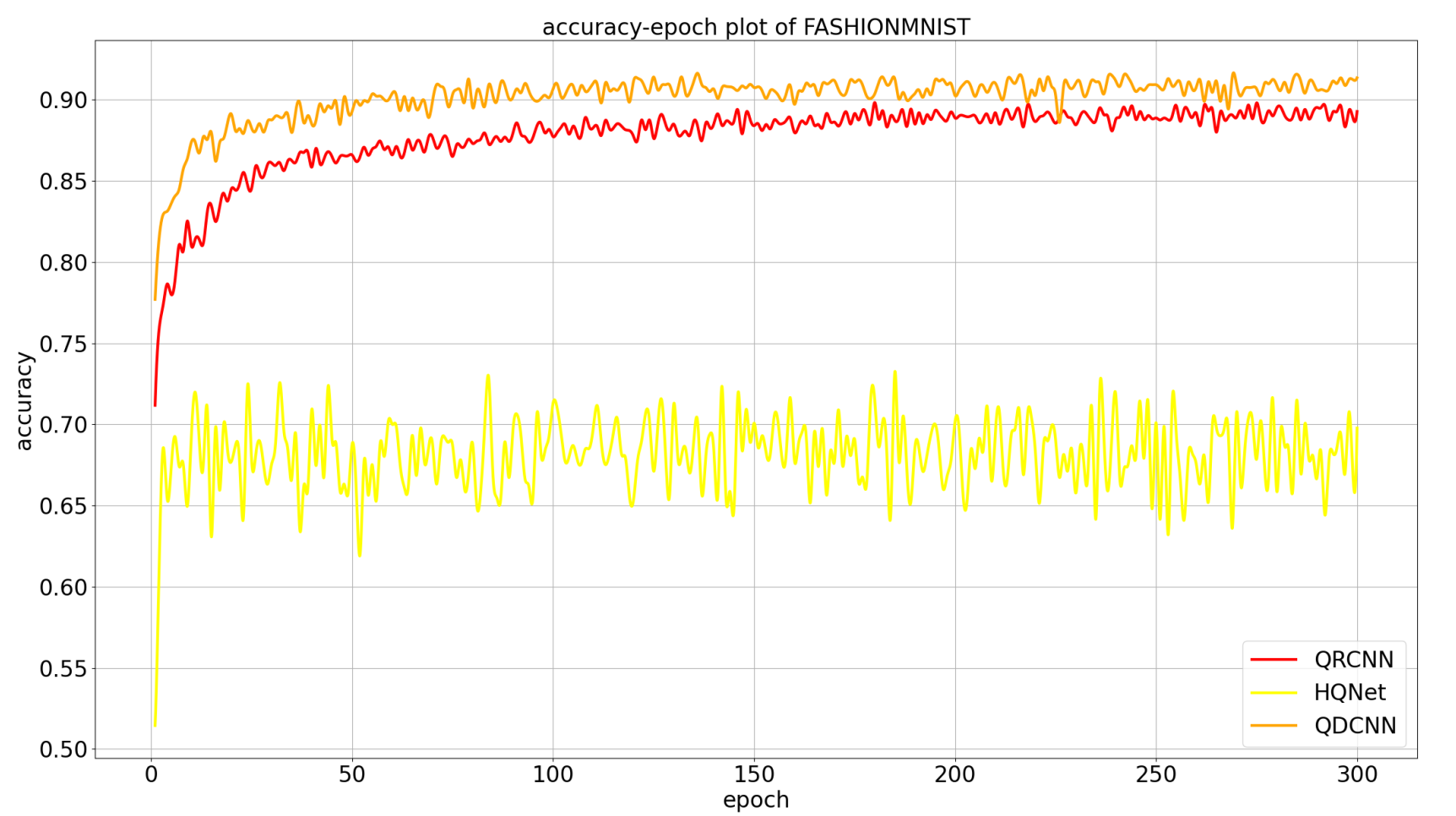}%
	}
	\hfil
	\subfloat[\centering]{\includegraphics[width=3.6in]{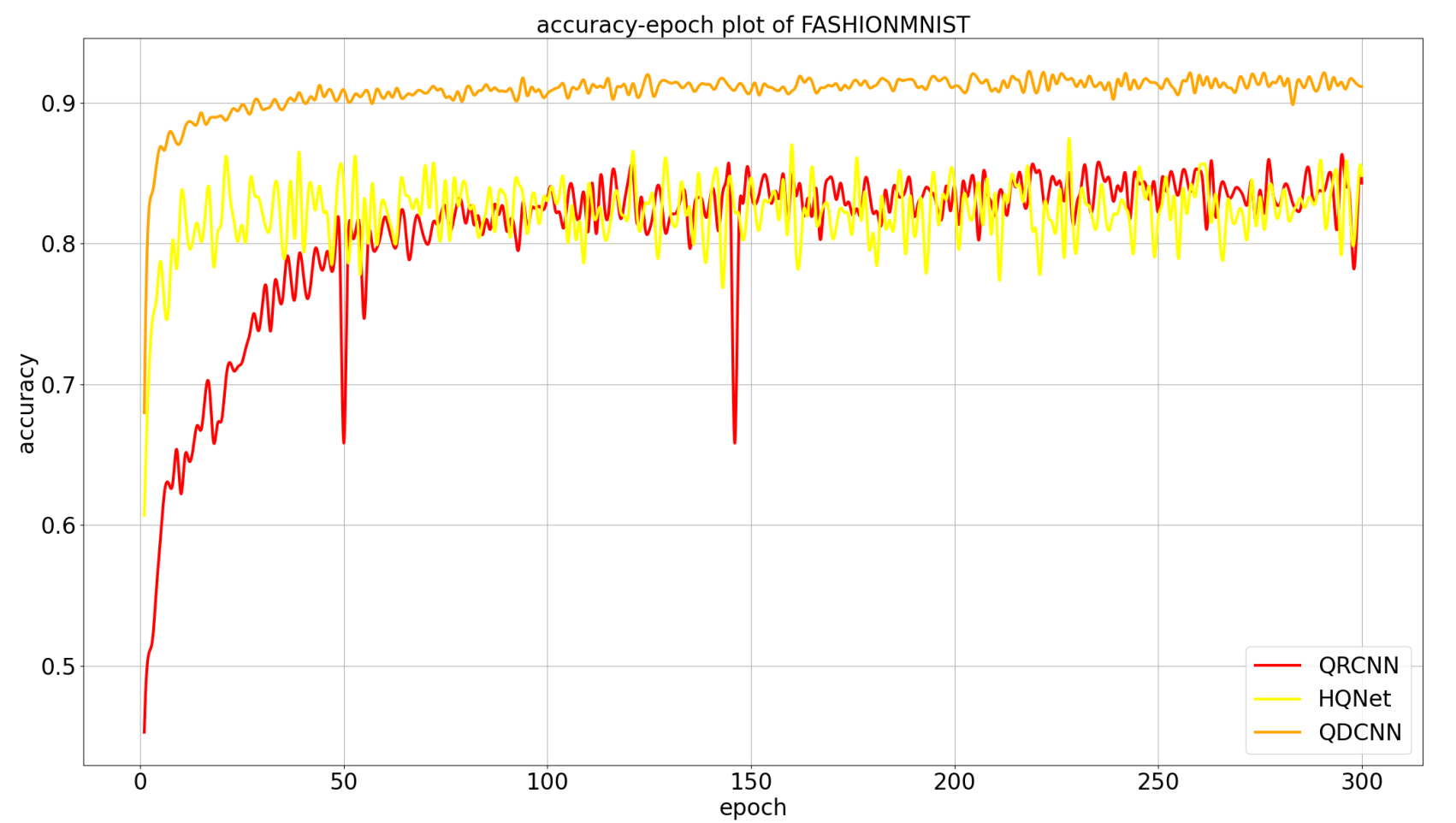}%
	}
	\caption{Accuracy results under symmetrical mixed noise attacks on FASHIONMNIST. We utilize a state-of-the-art HQNet for comparison. (a) Parameter attack in form \ding{172}. (b) Parameter attack in form \ding{173}. We can see that as for (a), the average accuracy of QDCNN is slightly higher than that of QRCNN and much higher than that of HQNet. In terms of (b), while the accuracy of HQNet and QRCNN is very close, that of QDCNN is higher than it.}
\end{figure*}
\begin{table}[!t]
	\centering
	\renewcommand{\arraystretch}{0.75}
	\caption{Accuracy on FASHIONMNIST under symmetrical noise attacking}
	\label{table_example}
	\centering
	\scalebox{0.8}{
		\begin{tabular}{c c c c c}
			\toprule
		noise type &  form	& QRCNN & HQNet & QDCNN \\
			\midrule
			mixed & \ding{172} & 0.8879$\pm$0.0102 & 0.6839$\pm$0.0132 & \textbf{0.9075$\pm$0.0096} \\
			mixed & \ding{173} & 0.8326$\pm$0.0116 & 0.8242$\pm$0.0117 & \textbf{0.9130$\pm$0.0094} \\
			gaussian & \ding{172} & 0.8634$\pm$0.0109 & 0.8792$\pm$0.0103 & \textbf{0.9109$\pm$0.0096} \\
			gaussian & \ding{173} & 0.8701$\pm$0.0108 & 0.8893$\pm$0.0102 & \textbf{0.9267$\pm$0.0091} \\
			uniform & \ding{172} & 0.8943$\pm$0.0105 & 0.8742$\pm$0.0110 & \textbf{0.9214$\pm$0.0091} \\
			uniform & \ding{173} & 0.9045$\pm$0.0100 & 0.9001$\pm$0.0099 & \textbf{0.9148$\pm$0.0088} \\
			\bottomrule
	\end{tabular}}
\end{table}
Fig.5 and Table 12 demonstrate the accuracy of HQNet, QRCNN and QDCNN on FASHIONMNIST under symmetrical attacks in the two forms. From the outcomes, QRCNN does not outperform HQNet, but QDCNN directy shows more advantages with the densely-connected architecture over \textcolor{black}{QRCNN} and HQNet. For instance, under mixed noise attacking in form \ding{173}, while the accuracy of QRCNN and HQNet is 83.26\%$\pm$1.16\%  and 82.42\%$\pm$1.17\%, that of QDCNN is 91.30\%$\pm$0.94\% , which is 8\%-9\% higher than that of HQNet and QRCNN. Comparison results between pure classical models and our hybrid models are in the section 8 of the supplementary material. Sometimes our hybrid models show slight more prominent performance than classical models, while the results of classical models are marginally better than that of our hybrid models at times. On the whole, our hybrid models indicate roughly the same level of capability as classical models in resistance to symmetrical noise attacks.
\subsubsection{Test under asymmetrical noise attacks}
\begin{figure*}[!t]
	\centering
	\subfloat[\centering]{\includegraphics[width=3.6in]{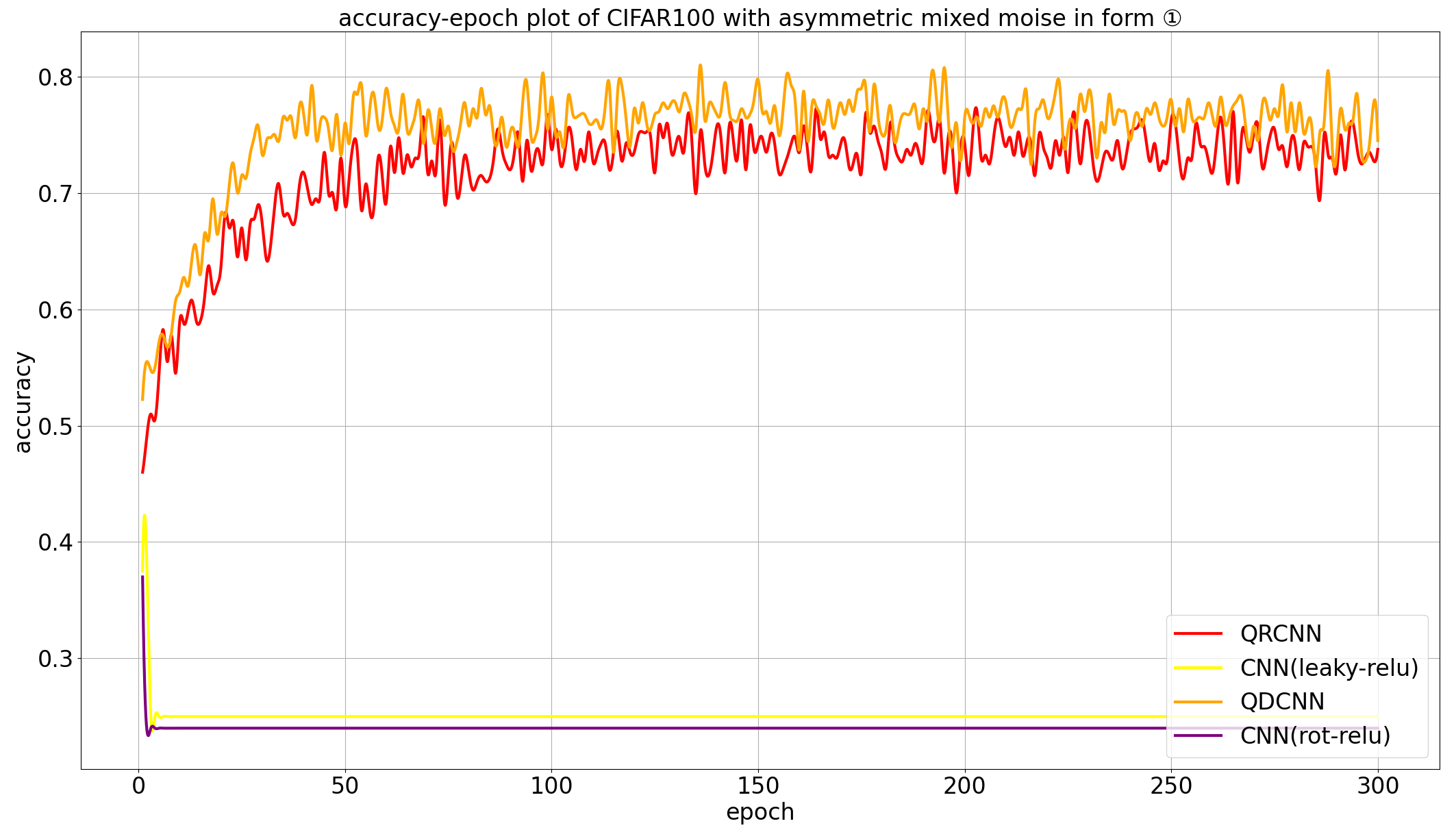}%
	}
	\hfil
	\subfloat[\centering]{\includegraphics[width=3.6in]{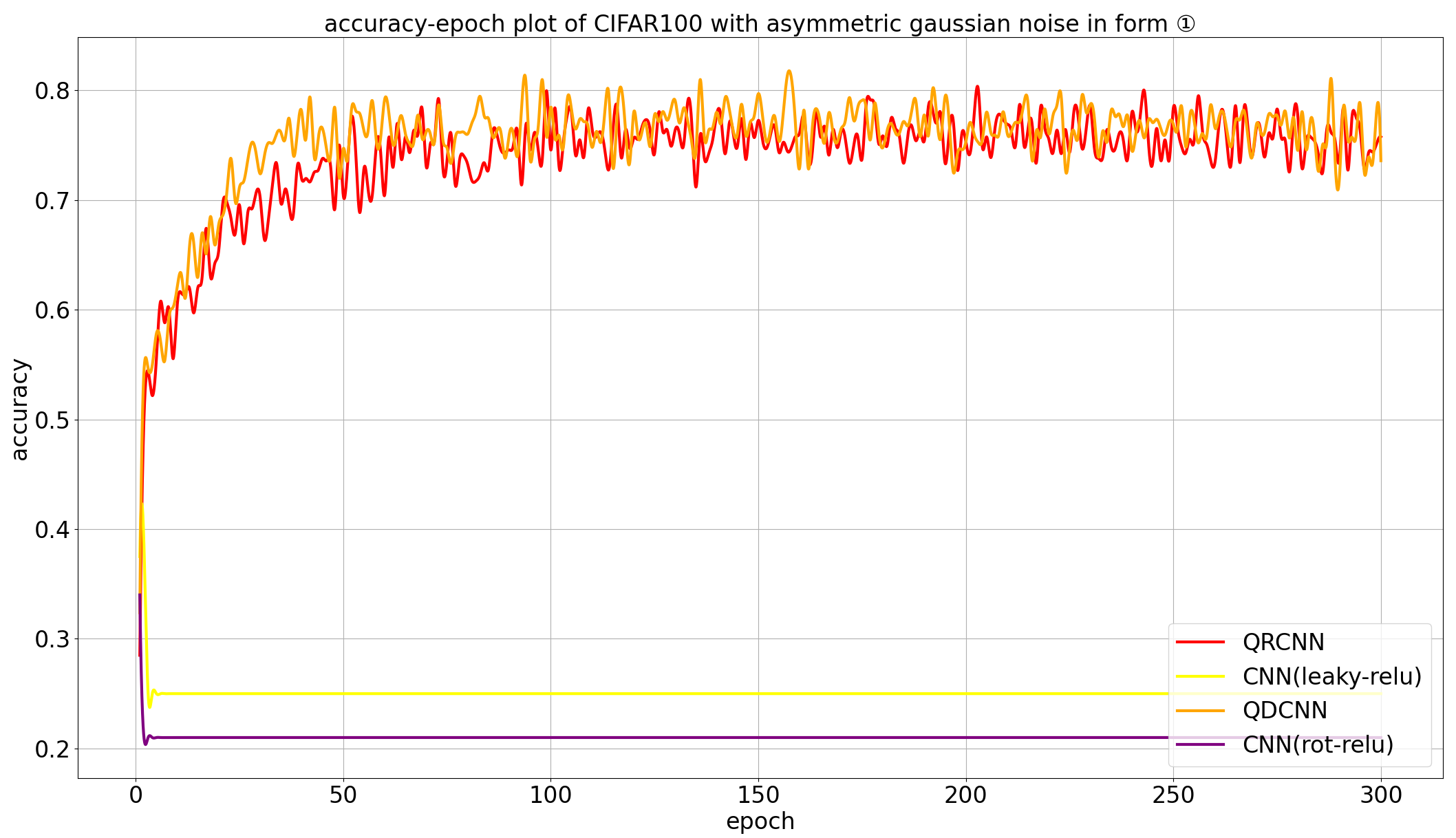}%
	}
	\caption{Asymmetrical noise attack in form \ding{172} in CIFAR100. We utilize two pure classical CNNs for comparison, of which the activation functions are separately rot-relu and leaky-relu functions. (a) Parameter attack with mixed asymmetrical noise. (b) Parameter attack with gaussian asymmetrical noise. In terms of the average test accuracy of the two pure classical models, the results remain low values in (a) and (b) since the loss reach nan.}
\end{figure*}
\begin{figure*}[!t]
	\centering
	\subfloat[\centering]{\includegraphics[width=3.6in]{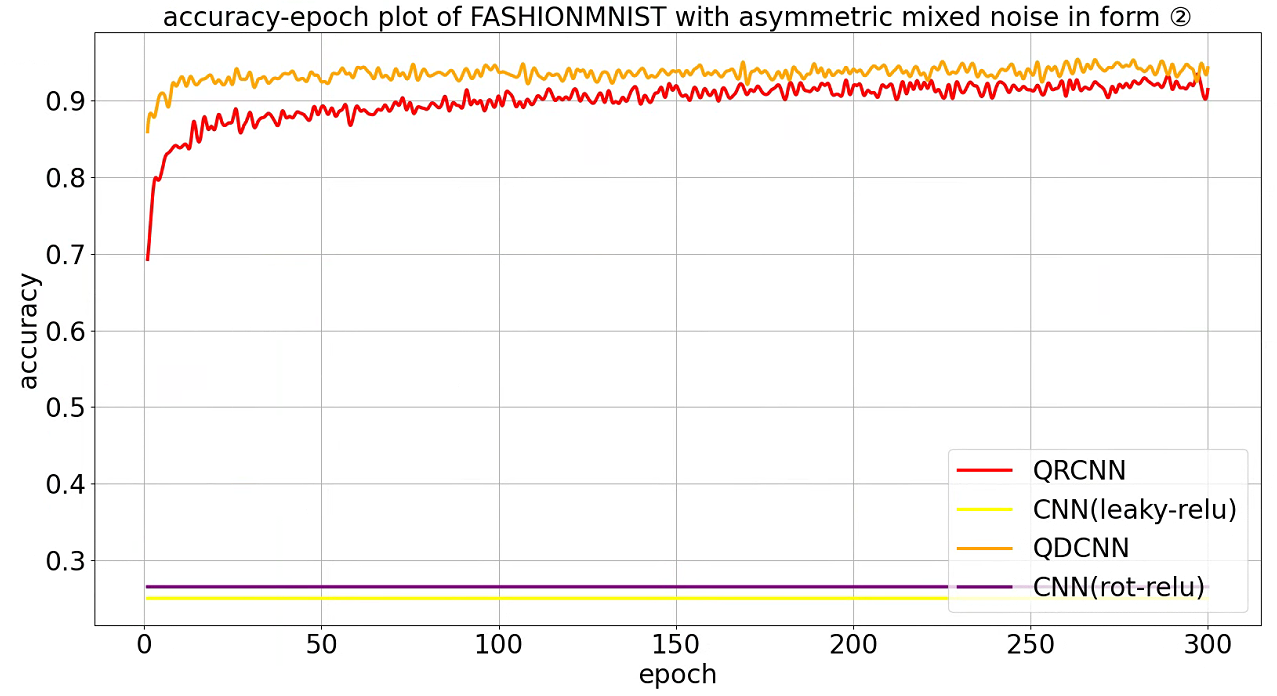}%
	}
	\hfil
	\subfloat[\centering]{\includegraphics[width=3.6in]{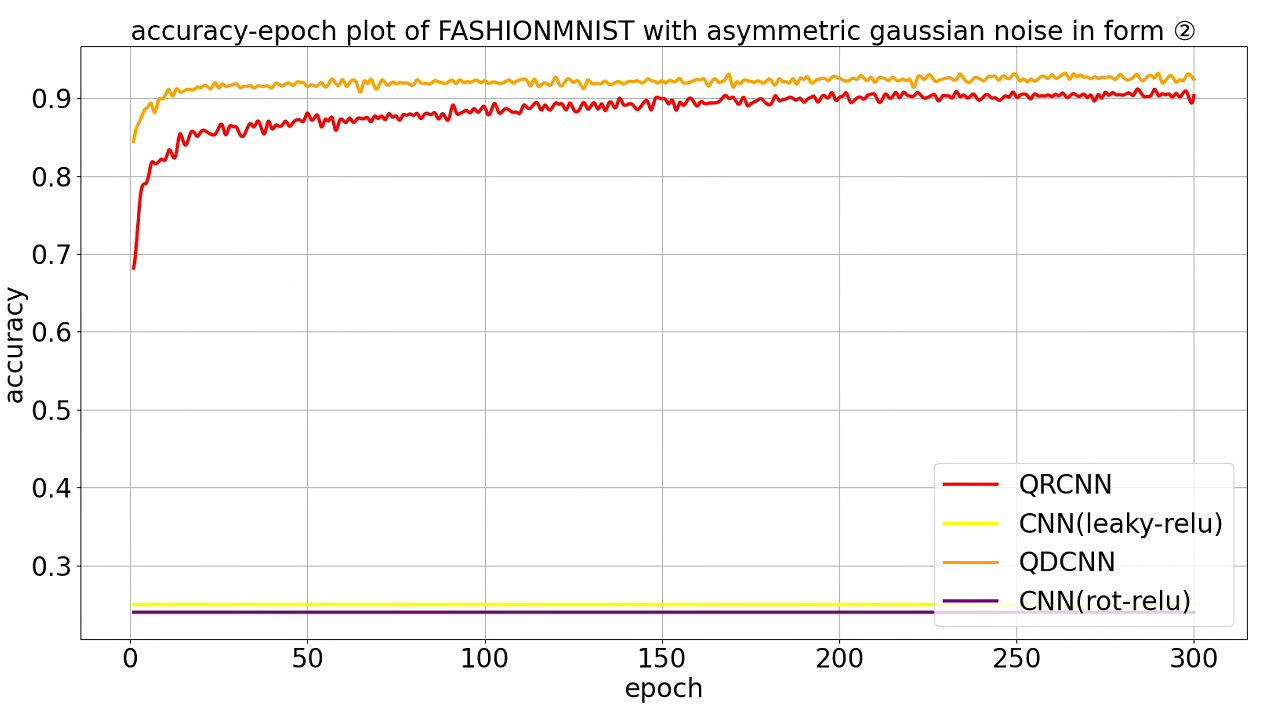}%
	}
	\caption{Accuracy under asymmetrical noise attack in form \ding{173} on FASHIONMNIST. We utilize two pure classical CNNs for comparison, of which the activation functions are separately rot-relu and leaky-relu function. (a) Parameter attack with mixed asymmetrical noise. (b) Parameter attack with gaussian asymmetrical noise. In terms of the average test accuracy of the two pure classical models, the results remain low values in (a) and (b) since the loss reach nan.}
\end{figure*}
\begin{figure*}[!t]
	\centering
	\includegraphics[width=5.5in]{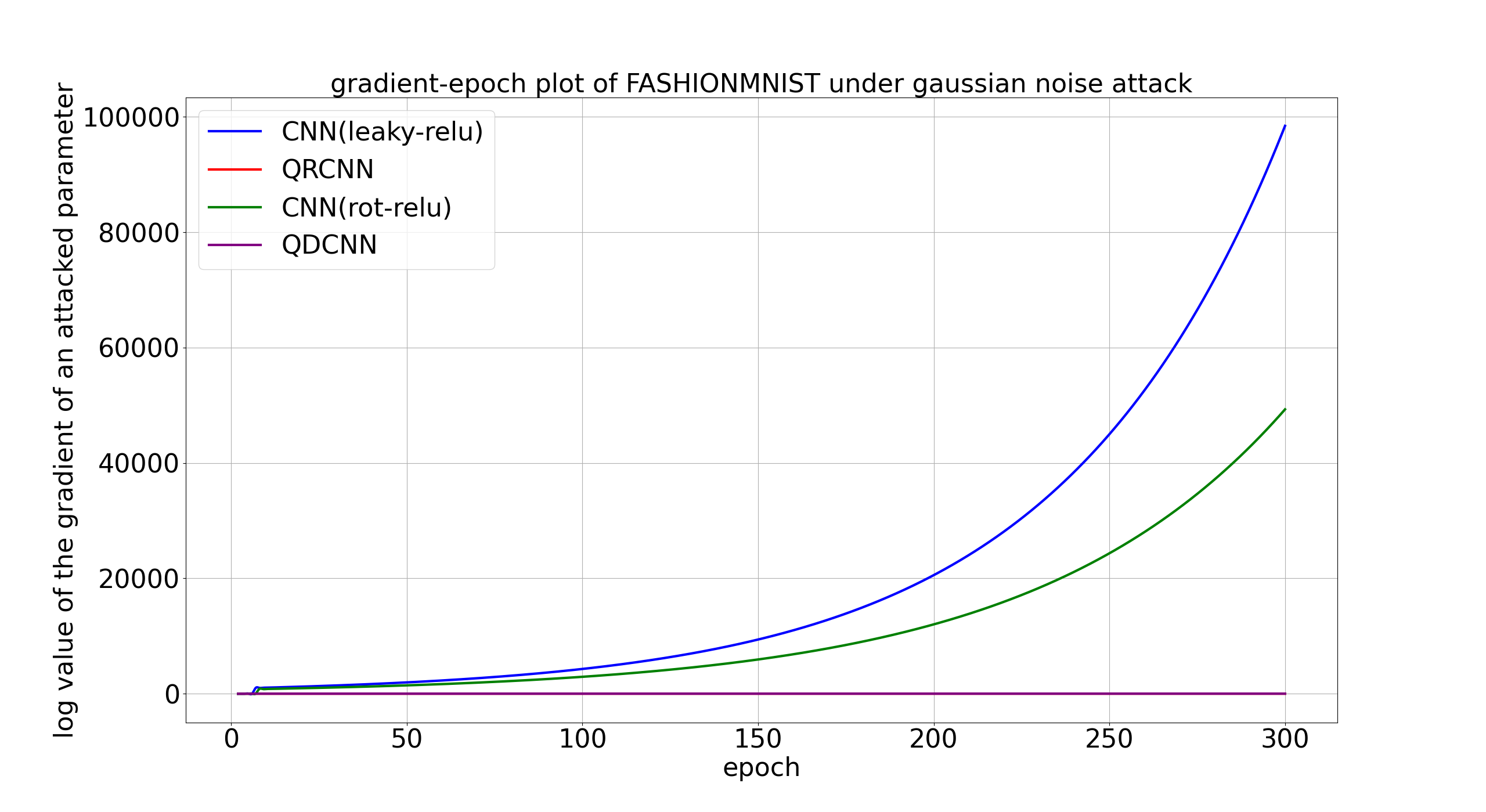}
	\caption{Log values of gradient of an attacked parameter. The noise utilized is asymmetrical gaussian one. As the graph shows, since gradient explosion occurs in two CNNs, their gradient log values increase exponentially, while those of QRCNN and QDCNN are close to 0.}
\end{figure*}
\begin{figure*}[!t]
	\centering
	\subfloat[\centering]{\includegraphics[width=3.6in]{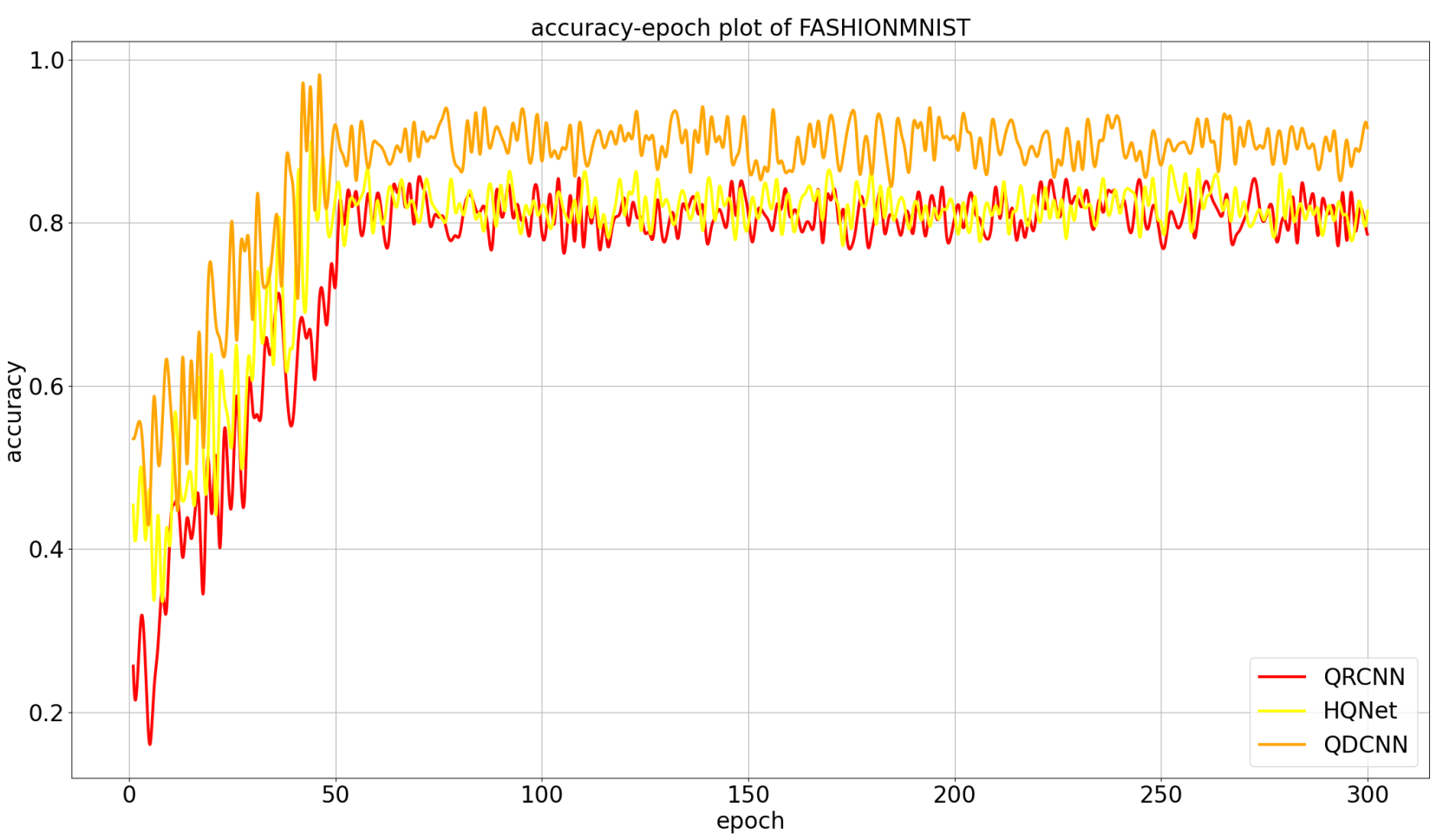}%
	}
	\hfil
	\subfloat[\centering]{\includegraphics[width=3.6in]{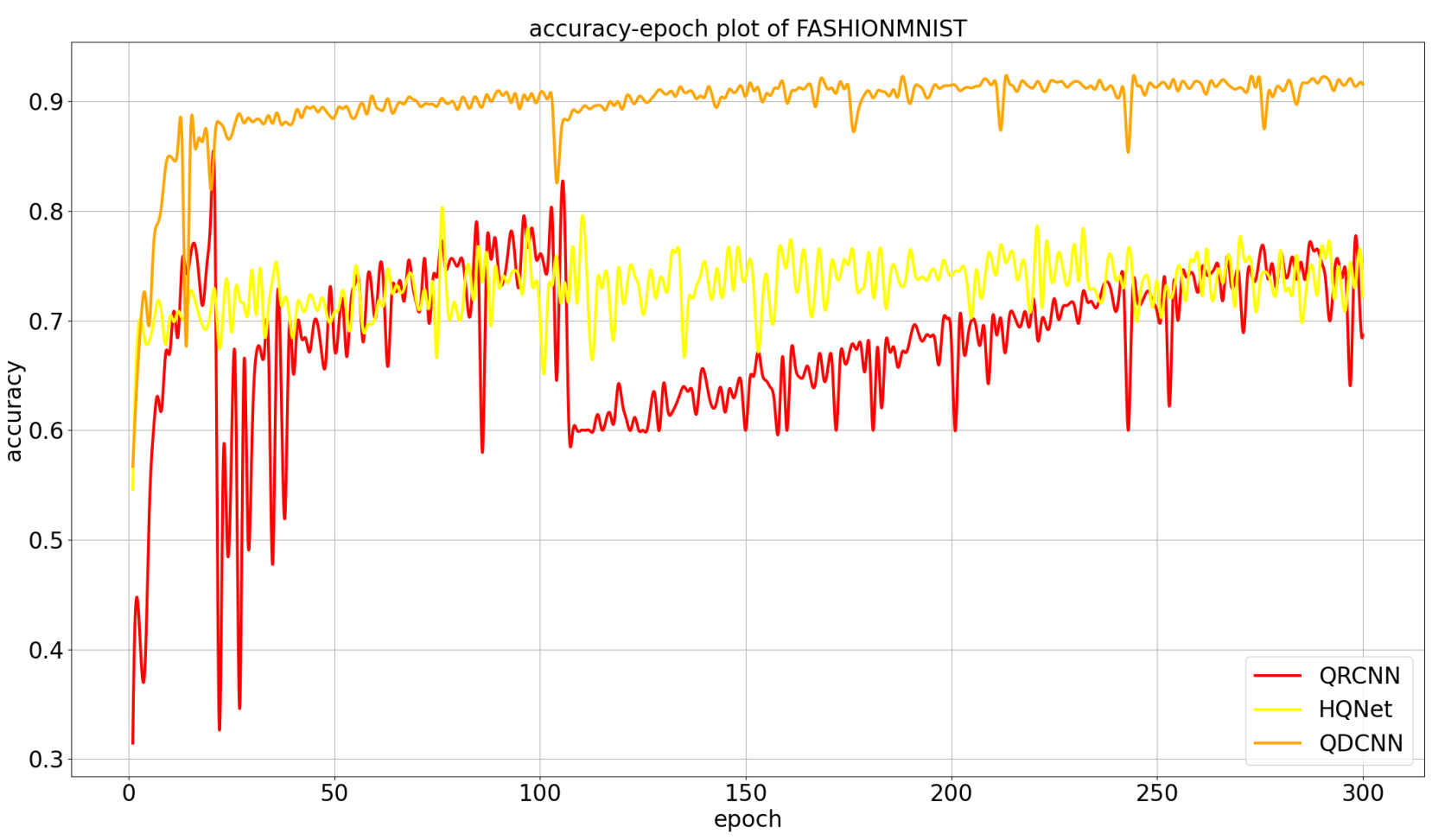}%
	}
	\caption{Accuracy under asymmetrical mixed noise attacks on FASHIONMNIST. We utilize a state-of-the-art HQNet for comparison. (a) Parameter attack in form \ding{172}. (b) Parameter attack in form \ding{173}.}
\end{figure*}

\begin{table*}[!t]
	\centering
	\renewcommand{\arraystretch}{0.8}
	\caption{Accuracy on FASHIONMNIST under asymmetrical noise attacking}
	\label{table_example}
	\centering
	\scalebox{0.8}{
		\begin{tabular}{c c c c c c}
			\toprule
			& & CNN(rot-relu)& QRCNN & CNN(leaky-relu) & QDCNN \\
			\midrule
			\multicolumn{1}{c|}{\multirow{2}{*}{Fig.(6a)}} & accuracy & 23.00\%$\pm$1.43\%  & 73.65\%$\pm$1.09\%  & 25.00\%$\pm$1.45\%  & \textbf{78.67\%$\pm$1.01\%} \\
			\multicolumn{1}{c|}{} & loss & nan & 0.0085 & nan & \textbf{0.0082} \\
			\hline
			\multicolumn{1}{c|}{\multirow{2}{*}{Fig.(6b)}} & accuracy & 21.00\%$\pm$1.42\%  & 75.96\%$\pm$1.05\%  & 25.00\%$\pm$1.45\%  & \textbf{76.10\%$\pm$1.04\%} \\
			\multicolumn{1}{c|}{} & loss & nan & \textbf{0.0063} & nan & 0.0079 \\
			\hline
			\multicolumn{1}{c|}{\multirow{2}{*}{Fig.(7a)}} & accuracy & 27.00\%$\pm$1.44\%  & 91.86\%$\pm$0.89\% & 25.00\%$\pm$1.43\% & \textbf{93.02\%$\pm$0.84\%} \\
			\multicolumn{1}{c|}{} & loss & nan & \textbf{0.0012} & nan & 0.0027 \\
			\hline
			\multicolumn{1}{c|}{\multirow{2}{*}{Fig.(7b)}} & accuracy & 24.00\%$\pm$1.43\% & 89.71\%$\pm$0.93\% & 25.00\%$\pm$1.43\% & \textbf{92.94\%$\pm$0.90\%} \\
			\multicolumn{1}{c|}{} & loss & nan & \textbf{0.0014} & nan & 0.0018 \\
			\bottomrule
	\end{tabular}}
\end{table*}

\begin{table}[!t]
	\centering
	\renewcommand{\arraystretch}{0.75}
	\caption{Accuracy on FASHIONMNIST under three asymmetrical noise attacks}
	\label{table_example}
	\centering
	\scalebox{0.8}{
		\begin{tabular}{c c c c c}
			\toprule
			noise type &  form	& QRCNN & HQNet & QDCNN \\
			\midrule
			mixed & \ding{172} & 0.8278$\pm$0.0107 & 0.8401$\pm$0.0102 & \textbf{0.8999$\pm$0.0092} \\
			mixed & \ding{173} & 0.6833$\pm$0.0122 & 0.7380$\pm$0.0117 & \textbf{0.9187$\pm$0.0087} \\
			gaussian & \ding{172} & 0.7993$\pm$0.0123 & 0.7612$\pm$0.0128 & \textbf{0.9056$\pm$0.0091} \\
			gaussian & \ding{173} & 0.8596$\pm$0.0099 & 0.8478$\pm$0.0100 & \textbf{0.9125$\pm$0.0088} \\
			uniform & \ding{172} & 0.7934$\pm$0.0122 & 0.8150$\pm$0.0111 & \textbf{0.9073$\pm$0.0095} \\
			uniform & \ding{173} & 0.8092$\pm$0.0110 & 0.8445$\pm$0.0116 & \textbf{0.9177$\pm$0.0087} \\
			\bottomrule
	\end{tabular}}
\end{table}
More importantly, our hybrid models show much greater superiority over classical models under asymmetrical noise attacks in the two forms (see supplementary materials section 8 for more results). Fig.6 shows the accuracy under gaussian and mixed asymmetrical attack in form \ding{172} on CIFAR100. As for Fig.(6a), while the accuracy of two CNNs is only 25\%$\pm$1.45\% and 23\%$\pm$1.43\% separately, that of QDCNN and QRCNN is 78.67\%$\pm$1.23\% and 73.65\%$\pm$1.29\% respectively, which shows the advantages of the quantum-inspired layers of our hybrid architectures. And there are also a little more superiority of QDCNN over QRCNN. The accuracy results of the four models in Fig.(6b) is similar with that in Fig.(6a). Fig.7 demonstrates the accuracy under gaussian and mixed asymmetrical attacks in form \ding{173} on FASHIONMNIST. In the two figures, the accuracy of the pure classical models maintains at a low value around 25\%, which shows that the pure classical models fail to learn the features. And we find the loss curves of the pure classical models oscillate badly and the loss values are very large and up to nan, which implies the reason of the bad performance of the pure classical models is gradient explosion (see section 9 of the supplementary material). However, even under various asymmetrical noise attacks with the two forms, our hybrid models still show great performance on the tasks with the accuracy being 90\% roughly due to the quantum-inspired layers in our hybrid models to avoid gradient explosion. And Table 13 shows the average accuracy values of the curves in Fig.6 and Fig.7.

\textcolor{black}{To further validate whether the reason of the terrible performance of classical models is from gradient explosion, we print the logarithm values of the gradient of the loss function to an attacked parameter of our hybrid models and classical CNNs. We find that the gradient values of CNNs grow rapidly and reach nan with epoch increasing. Hence, we have to print logarithm values of the gradient of an attacked parameter of CNNs. As is shown in Fig.8, while the curves of QRCNN and QDCNN are flat, of which the values are close to 0, those of two CNNs rise rapidly, which is consistent with the characteristics of gradient explosion.}

Moreover, Fig.9 as well as Table 14 indicate the accuracy of QRCNN, QDCNN and HQNet under assorted asymmetrical noises attacks with the two forms on FASHIONMNIST. Both Fig.(9a) and (9b) shows the accuracy of the three models with that of QDCNN greater than HQNet and QRCNN, which points the advantages of the combination of dense connection with the convolutional layers and the symmetrical circuit model. In Fig.(9b), the fluctuation degree of QRCNN is slightly higher than that of QDCNN and HQNet, which shows the instability of QRCNN. However, the average accuracy of QRCNN and HQNet in Table 14 is close, which shows the same level of performance of them. 

\begin{figure*}[!t]
	\centering
	\subfloat[\centering]{\includegraphics[width=3.6in]{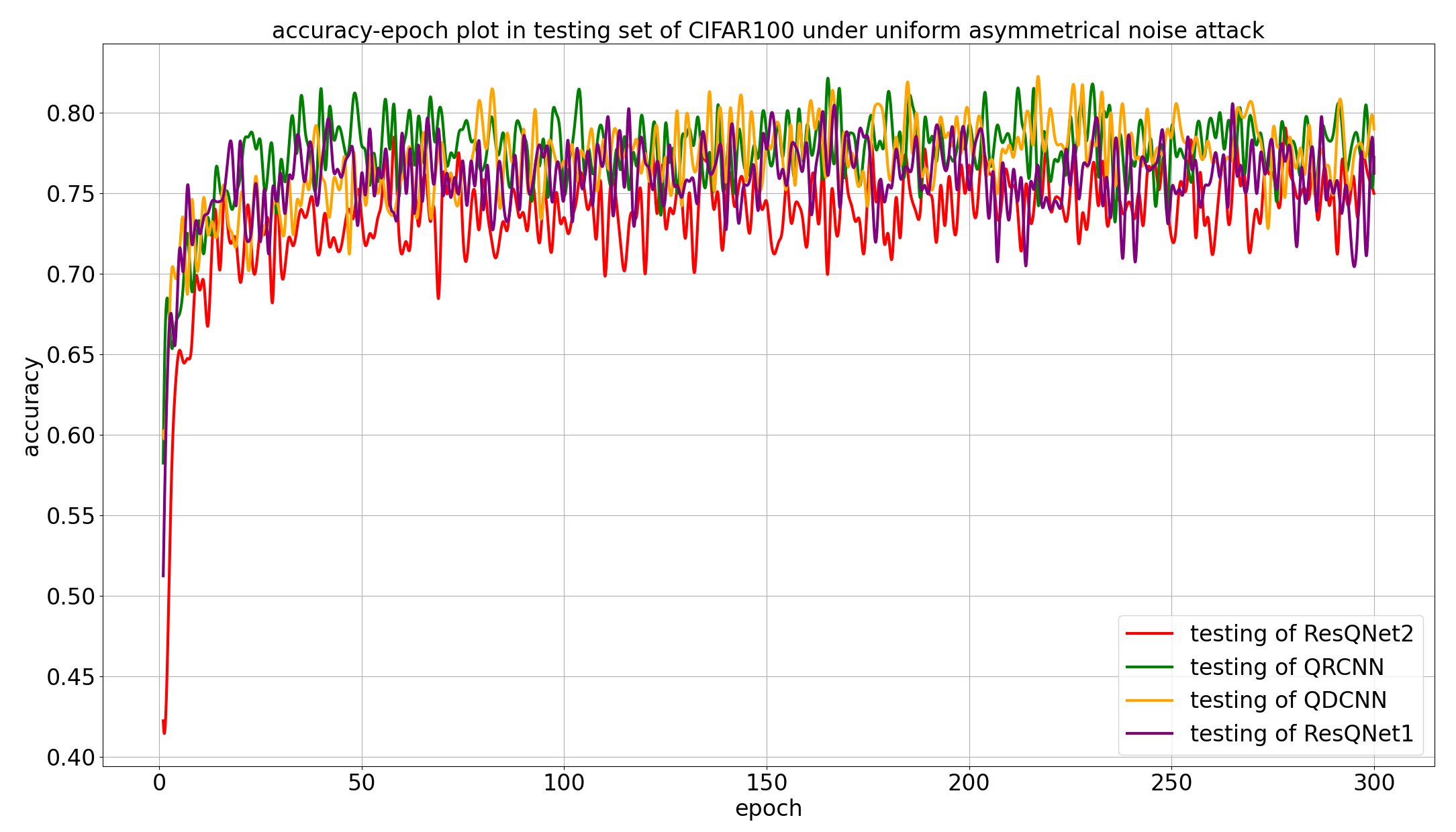}%
	}
	\hfil
	\subfloat[\centering]{\includegraphics[width=3.6in]{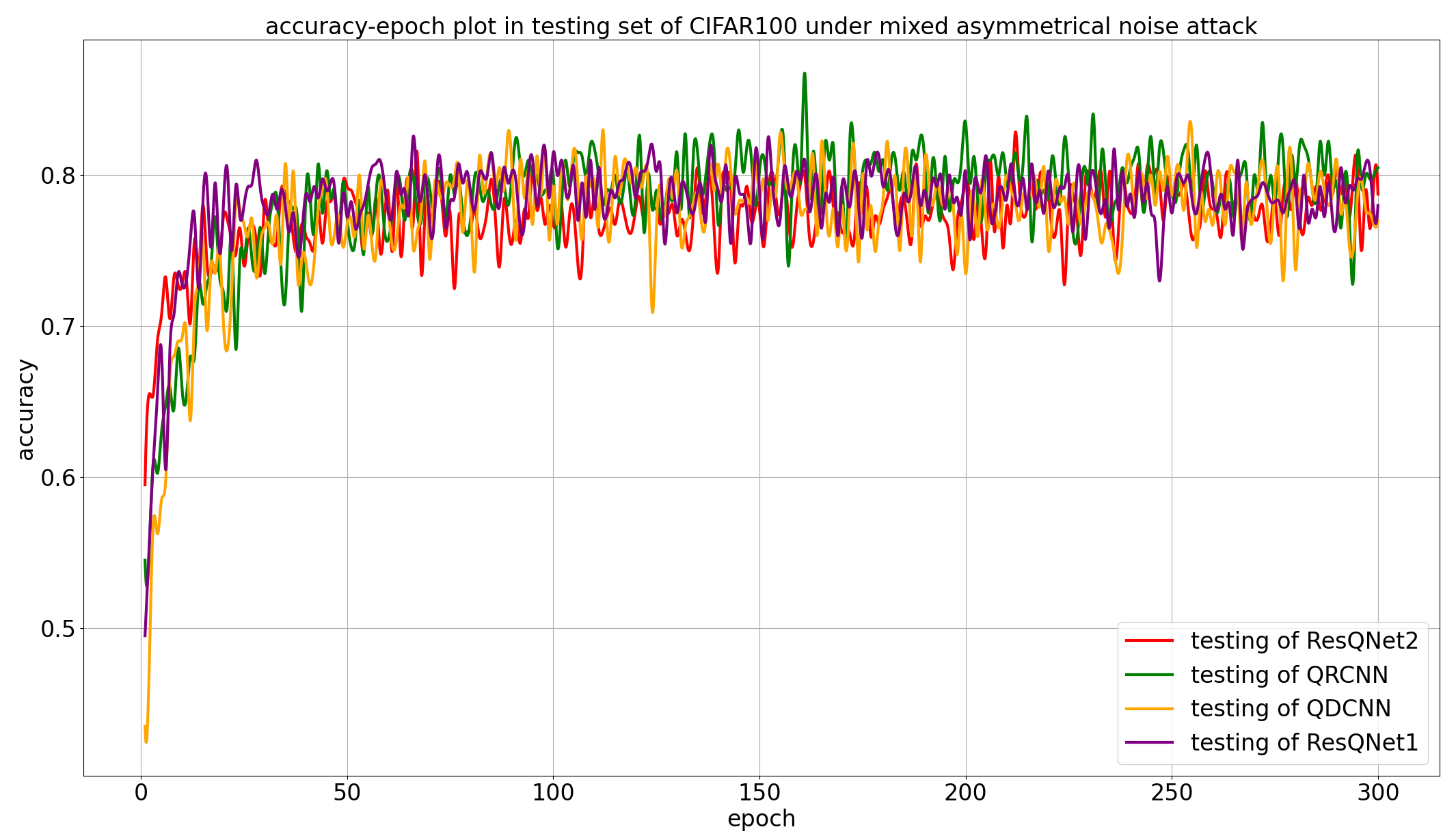}%
	}
	\caption{Accuracy under asymmetrical noise attacks on CIFAR10 in form \ding{172}. We utilize ResQNet1 and ResQNet2 for comparison. (a) uniform noise attack (b) mixed noise attack.}
\end{figure*}

\begin{figure*}[!t]
	\centering
	\subfloat[\centering]{\includegraphics[width=3.6in]{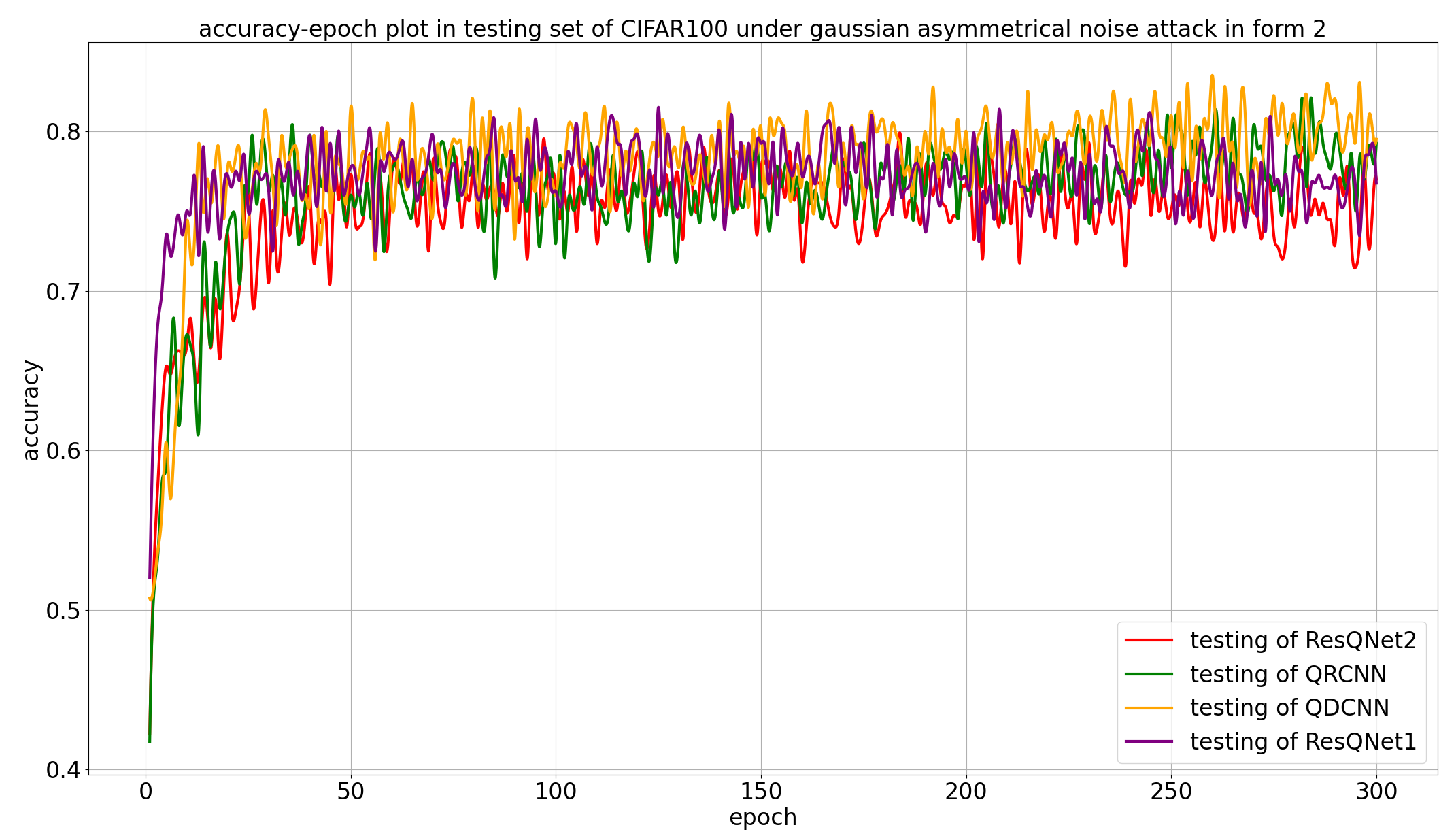}%
	}
	\hfil
	\subfloat[\centering]{\includegraphics[width=3.6in]{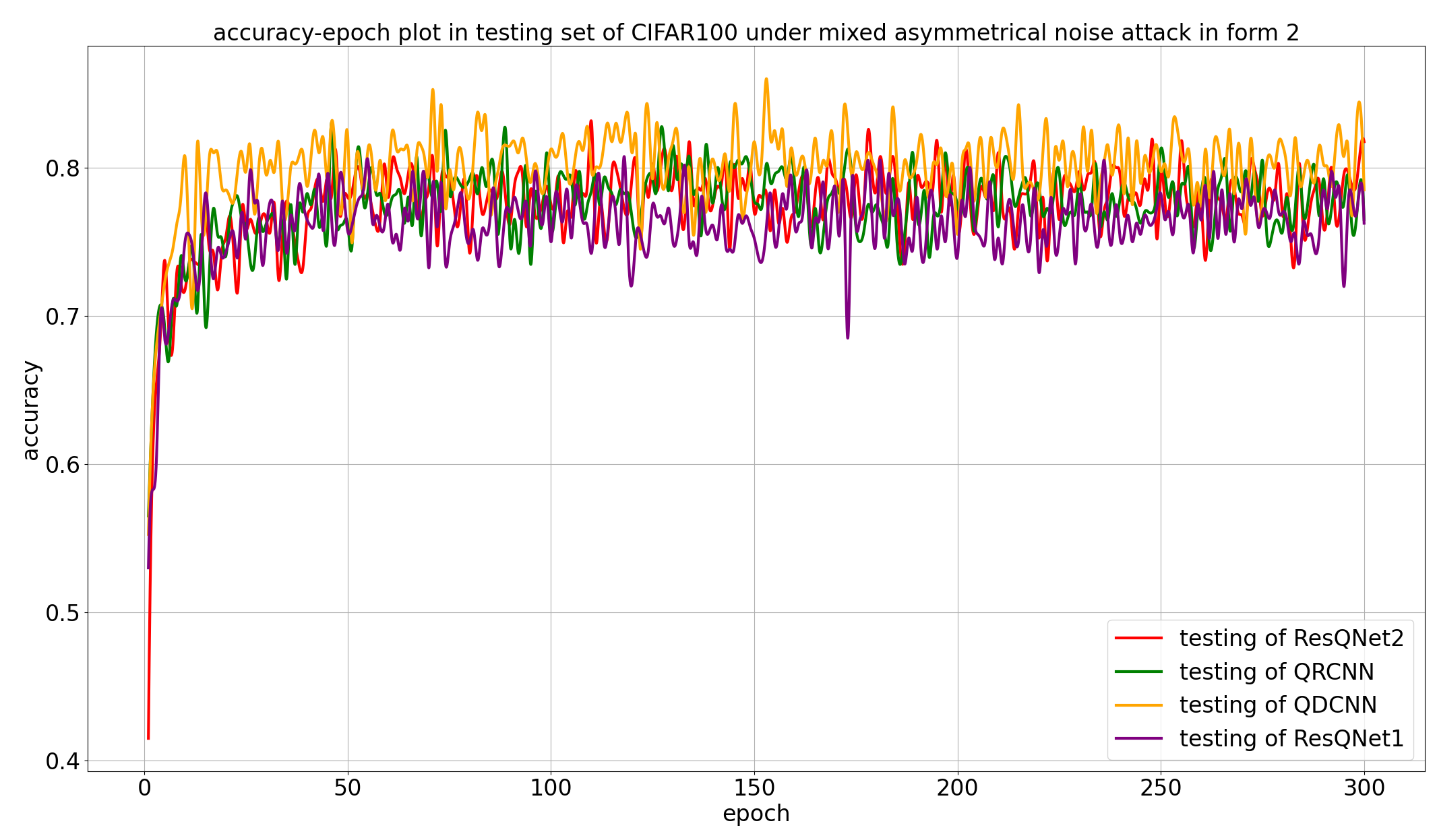}%
	}
	\caption{Accuracy under asymmetrical noise attacks on CIFAR10 in form \ding{173}. We utilize ResQNet1 and ResQNet2 for comparison. (a) gaussian noise attack (b) mixed noise attack.}
\end{figure*}

\begin{table*}[!t]
	\centering
	\renewcommand{\arraystretch}{0.75}
	\caption{Accuracy on CIFAR10 under three asymmetrical noises attack in form \ding{172} and \ding{173}}
	\label{table_example}
	\centering
	\scalebox{0.8}{
		\begin{tabular}{c c c c c c}
			\toprule
		attack form &noise type&  ResQNet1&ResQNet2 & QRCNN & QDCNN  \\
			\midrule
         \ding{172} & uniform & 76.21\%$\pm$2.02\%&74.55\%$\pm$2.07\% & \textbf{77.95\%$\pm$1.97\%} & 77.92\%$\pm$1.97\%  \\
         \ding{172} & mixed & 78.80\%$\pm$1.94\%&77.84\%$\pm$1.97\% & \textbf{79.81\%$\pm$1.91\%} & 78.80\%$\pm$1.95\%  \\
		 \ding{173} & mixed & 76.47\%$\pm$2.02\%&78.20\%$\pm$1.96\% & 78.07\%$\pm$1.97\% & \textbf{80.31\%$\pm$1.89\%}  \\
		 \ding{173} & gaussian & 77.41\%$\pm$1.99\%&75.79\%$\pm$2.03\% & 77.20\%$\pm$1.99\% & \textbf{79.20\%$\pm$1.93\%}  \\
		    \bottomrule
	\end{tabular}}
\end{table*}
\textcolor{black}{Besides that, Fig.10 shows the accuracy curves of ResQNet1, ResQNet2 and our hybrid models to three asymmetrical noise attacks in form \ding{172} on CIFAR10. And Fig.11 shows that of ResQNet1, ResQNet2 and our hybrid models to three asymmetrical noise attacks in form \ding{173} on CIFAR10. All the curves in both the figures firstly rise and then fluctuate around certain values. Also, Table 15 showcases the exact average values of accuracy of the four models with standard deviation in Fig.10 and Fig.11, from which we can see QRCNN and QDCNN slightly outperform ResQNet1 and ResQNet2. For instance, shown on Table 15, under gaussian noise attacks in form \ding{173}, while the accuracy of ResQNet1 and ResQNet2 on CIFAR10 is separately 77.41\% and 75.79\%, that of QRCNN and QDCNN is respectively 77.20\% and 79.20\%. It also tells the advantage of the residual and dense mappings in our hybrid models. It is also valuable to explore the combination of circuits and other residual and dense connections to further facilitate the accuracy of hybrid neural networks.} 
\subsubsection{Ablation study}
To test the advantages of the adaptive residual and dense connection of our hybrid models, we use \textbf{T}raditional \textbf{Q}uantum-inspired \textbf{F}eedforward \textbf{N}eural \textbf{N}etwork (TQFNN) and \textbf{T}raditional \textbf{Q}uantum-inspired \textbf{C}onvolutional \textbf{N}eural \textbf{N}etwork (TQCNN) for comparison. The framework of TQFNN is exactly the same as that of QRFNN and QDFNN but without residual or dense connection. Similarly, the architecture of TQCNN is also identical to that of QRCNN and QDCNN but without residual or dense connection in the convolutional and fully-connected layers as well. Hence, through theoretical analysis in the section 4 of the supplementary material, the TQFNN, QRFNN and QDFNN models are close in the computational and parameter complexity. Moreover, experimental results demonstrate that the accuracy of our hybrid models is approximately 2\%-3\% higher than that of TQFNN and TQCNN. For example, as for Fig.(12a), the accuracy of TQFNN on clean iris data is 87.93\%$\pm$1.04\%, while that of QRFNN and QDFNN is 89.57\%$\pm$0.98\% and 92.55\%$\pm$0.93\%. It showcases the superiority of the adaptive residual and dense connections in our hybrid models over the quantum-inspired neural networks without residual or dense connection. Furthermore, QDFNN and QDCNN outperform QRFNN and QRCNN a little separately, which shows the more advantages of the dense connection in QDFNN and QDCNN over the residual connection in QRFNN and QRCNN.
\begin{figure*}[!t]
	\centering
	\subfloat[\centering]{\includegraphics[width=3.6in]{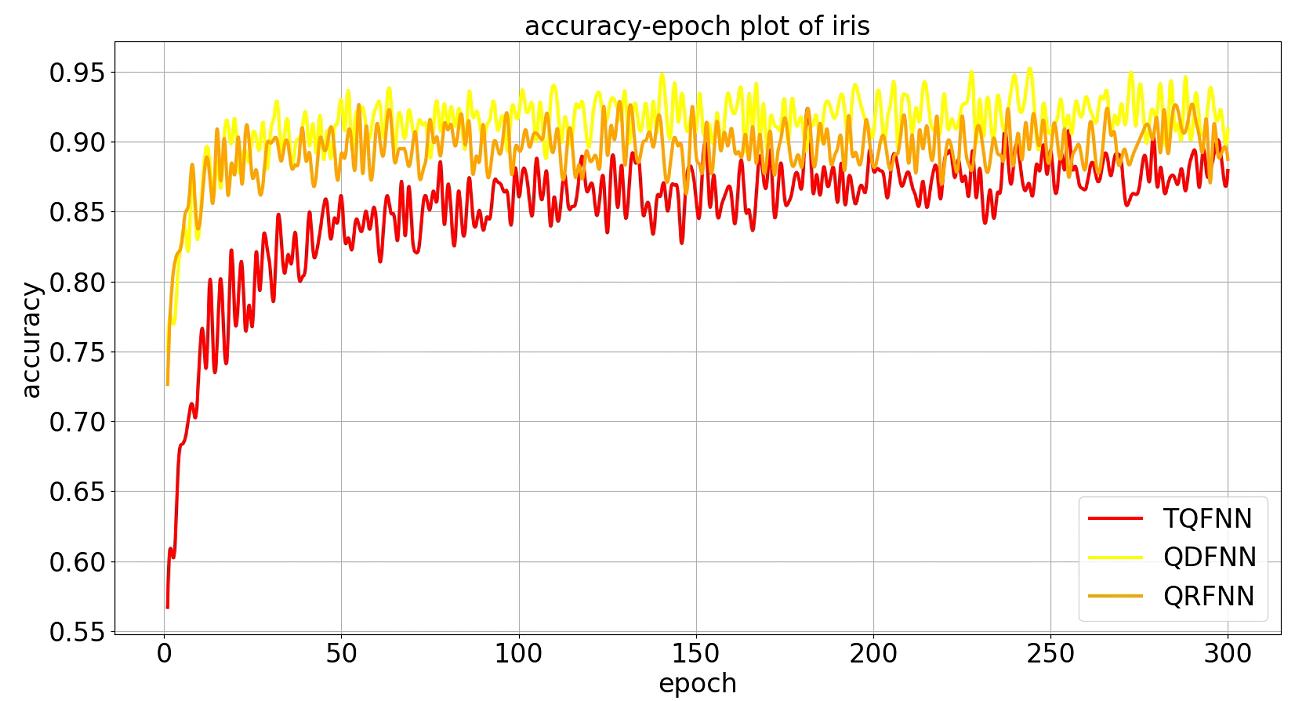}%
	}
	\hfil
	\subfloat[\centering]{\includegraphics[width=3.6in]{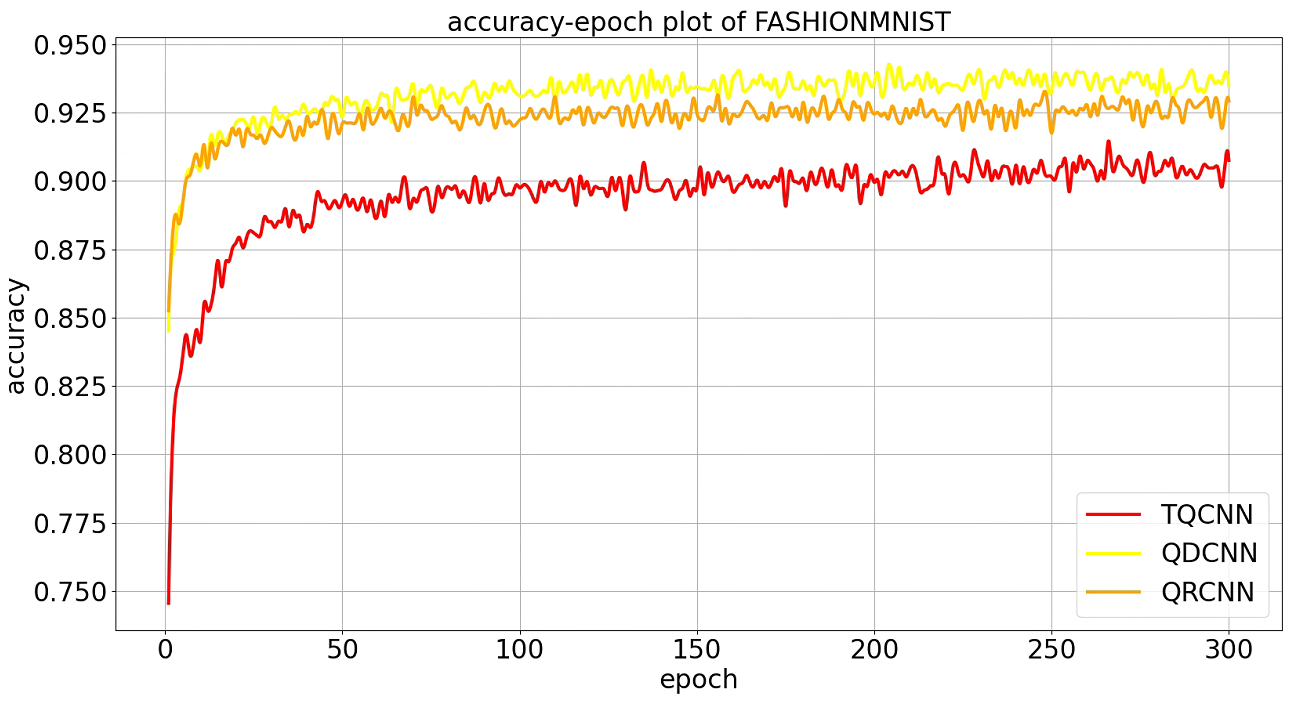}%
	}
	\caption{Test accuracy of the traditional quantum-inspired neural network and our hybrid models with noiseless datasets. (a) Accuracy results of noiseless iris data of TQFNN, QRFNN and QDFNN. (b) Accuracy results of noiseless FASHIONMNIST data of TQCNN, QRCNN and QDCNN. The average accuracy of TQCNN is 90.19\%$\pm$0.95\%, while that of QRCNN and QDCNN is 92.52\%$\pm$0.91\% and 93.55\%$\pm$0.86\%. It indicates the slightly more advantage of the dense connection in QDCNN over the residual connection in QRCNN. Additionally, our two hybrid models outperform TQCNN, which shows the function of the residual and dense connections.}
\end{figure*}
\section{Conclusion and discussion}
To summarize, we have firstly proposed the hybrid quantum-inspired neural networks amalgamating symmetrical circuits with adaptive residual or dense connections. We explain their frameworks and assess their generalization power and robustness. ResQNet1, ResQNet2, HQNet and classical MLP, CNNs with concrete structures and traditional quantum-inspired neural networks without residual or dense connection are for comparison. The charm of our hybrid models lies in these facts that:
\begin{itemize}
	\item{under the premise that the computational and parameter complexity of our hybrid models and the traditional quantum-inspired network without residual or dense connection are close, our hybrid models show 2\%-3\% higher accuracy over the quantum-inspired neural networks;}
	\item{they show the same level of generalization ability and robustness as the MLPs and CNNs when the datasets contain the six noises, or the parameters are attacked by symmetrical noises;}
	\item{they show much more remarkable robustness than the MLPs and CNNs with the parameters attacked by asymmetrical noises;}
	\item{the densely-connected frameworks possess slightly more superiority over HQNet on classification tasks with noisy datasets used;}
	\item{the densely-connected frameworks are a bit more advantageous than HQNet on classification tasks under the attacks of symmetrical and asymmetrical noises;}
	\item{\textcolor{black}{their frameworks are a little bit more superior than ResQNet1 and ResQNet2 on classification tasks with asymmetrically noisy datasets or to asymmetrical noise attacks;}}
	\item{They raise great potential to systematically prevent the gradient explosion problem.}
\end{itemize}

While the quantum-inspired layers enhance the robustness and prevent gradient explosion, the adaptive residual and dense connections facilitate feature learning and generalization power of our hybrid models. Therefore, the biggest novelty of our models lies in the combination of the proposed adaptive residual or dense connection with proposed symmetrical circuits, which improves the comprehensive performance in the neural network field. Hence, our hybrid models are able to substitute pure classical or other quantum-inspired neural networks in certain tasks. However, according to the section 5 of the supplementary material, the computational complexity of QRFNN, $C_{QRFNN} \approx B \cdot I \cdot L \cdot N^4 \propto N^4$, and the computational complexity of QDFNN, $C_{QDFNN} \approx B \cdot I \cdot L \cdot N^4 \propto N^4$, where $B$ and $I$ are batchsize and number of iterations in an epoch separately. Since $N = 2^n$, $C_{QRFNN} \propto 2^{4n}$, and $C_{QDFNN} \propto 2^{4n}$, where n is the number of qubit. It indicates the trouble to expand the dimension of the quantum state in the quantum-inspired layer, which is also the width of our hybrid neural networks \cite{19,48}. If our hybrid models are utilized for large-scale datasets, the computational complexity of them could increase exponentially, leading to a large amount of time consumption. Therefore, our models are more suitable for problems on data of modest-scale dimension of the feature space, such as iris data in scikit-learn library \cite{28}. Nonetheless, when it comes to the design of deep neural networks, our work may herald a future where the hybrid architectures redefine the boundaries of deep learning to unprecedented heights. What's more, owing to the outstanding generalization capability and robustness of our hybrid models, it makes sense to leverage them on complicated noisy engineering environments involving the coupled effects of multiple factors, such as unstructured data and noise interference \cite{44}. \textcolor{black}{To exemplify, it does make sense to evaluate the robustness of our hybrid models when there exists a substantial distributional discrepancy between training data and practical engineering scenarios, which would significantly facilitate the applicability of our models in real-world contexts.} The links of the supplementary material and program codes are below the acknowledgements. 
\section*{Acknowledgements}
This work is supported by the National Natural Science Foundation of China (Grants No. 12175104 and No. 12274223), the Innovation Program for Quantum Science and Technology (2021ZD0301701), the National Key Research and development Program of China (No. 2023YFC2205802), the Key Research and Development Program of Nanjing Jiangbei New Area (No.ZDYD20210101), the Program for Innovative Talents and Entrepreneurs in Jiangsu (No. JSSCRC2021484), and the Program of Song Shan Laboratory (Included in the management of Major Science and Technology Program of Henan Province) (No. 221100210800-02).

\begin{footnotesize}
	\href{https://drive.google.com/file/d/1Z6ZUUvyHvch_L2Gymg9Ol325pCu8_p8F/view?usp=sharing}{Click here for supplementary material information}
\end{footnotesize}\par

\begin{footnotesize}
	\href{https://drive.google.com/drive/folders/1n8mtdx5RWPLKnZvzjKZpHFTOHk0ZKpDP?usp=sharing}{Click here for program codes}
\end{footnotesize}

\bibliography{ref}

\begin{thebibliography}{52}
\expandafter\ifx\csname natexlab\endcsname\relax\def\natexlab#1{#1}\fi
\providecommand{\url}[1]{\texttt{#1}}
\providecommand{\href}[2]{#2}
\providecommand{\path}[1]{#1}
\providecommand{\DOIprefix}{doi:}
\providecommand{\ArXivprefix}{arXiv:}
\providecommand{\URLprefix}{URL: }
\providecommand{\Pubmedprefix}{pmid:}
\providecommand{\doi}[1]{\href{http://dx.doi.org/#1}{\path{#1}}}
\providecommand{\Pubmed}[1]{\href{pmid:#1}{\path{#1}}}
\providecommand{\bibinfo}[2]{#2}
\ifx\xfnm\relax \def\xfnm[#1]{\unskip,\space#1}\fi
%Type = Article
\bibitem[{LeCun et~al.(2015)LeCun, Bengio, and Hinton}]{1}
\bibinfo{author}{Y.~LeCun}, \bibinfo{author}{Y.~Bengio},
  \bibinfo{author}{G.~Hinton},
\newblock \bibinfo{title}{Deep learning},
\newblock \bibinfo{journal}{nature} \bibinfo{volume}{521}
  (\bibinfo{year}{2015}) \bibinfo{pages}{436--444}.
  \DOIprefix\doi{10.1038/nature14539}.
%Type = Book
\bibitem[{Goodfellow et~al.(2016)Goodfellow, Bengio, and Courville}]{2}
\bibinfo{author}{I.~Goodfellow}, \bibinfo{author}{Y.~Bengio},
  \bibinfo{author}{A.~Courville}, \bibinfo{title}{Deep learning},
  \bibinfo{publisher}{MIT press}, \bibinfo{year}{2016}.
%Type = Book
\bibitem[{Prince(2023)}]{3}
\bibinfo{author}{S.~J. Prince}, \bibinfo{title}{Understanding Deep Learning},
  \bibinfo{publisher}{MIT press}, \bibinfo{year}{2023}.
%Type = Article
\bibitem[{Schmidhuber(2015)}]{4}
\bibinfo{author}{J.~Schmidhuber},
\newblock \bibinfo{title}{Deep learning in neural networks: An overview},
\newblock \bibinfo{journal}{Neural networks} \bibinfo{volume}{61}
  (\bibinfo{year}{2015}) \bibinfo{pages}{85--117}.
  \DOIprefix\doi{10.1016/j.neunet.2014.09.003}.
%Type = Book
\bibitem[{Buduma et~al.(2022)Buduma, Buduma, and Papa}]{5}
\bibinfo{author}{N.~Buduma}, \bibinfo{author}{N.~Buduma},
  \bibinfo{author}{J.~Papa}, \bibinfo{title}{Fundamentals of deep learning},
  \bibinfo{publisher}{O'Reilly Media, Inc.}, \bibinfo{year}{2022}.
%Type = Article
\bibitem[{Guo et~al.(2016)Guo, Liu, Oerlemans, Lao, Wu, and Lew}]{6}
\bibinfo{author}{Y.~Guo}, \bibinfo{author}{Y.~Liu},
  \bibinfo{author}{A.~Oerlemans}, \bibinfo{author}{S.~Lao},
  \bibinfo{author}{S.~Wu}, \bibinfo{author}{M.~S. Lew},
\newblock \bibinfo{title}{Deep learning for visual understanding: A review},
\newblock \bibinfo{journal}{Neurocomputing} \bibinfo{volume}{187}
  (\bibinfo{year}{2016}) \bibinfo{pages}{27--48}.
  \DOIprefix\doi{10.1016/j.neucom.2015.09.116}.
%Type = Inproceedings
\bibitem[{Ngiam et~al.(2011)Ngiam, Khosla, Kim, Nam, Lee, and Ng}]{7}
\bibinfo{author}{J.~Ngiam}, \bibinfo{author}{A.~Khosla},
  \bibinfo{author}{M.~Kim}, \bibinfo{author}{J.~Nam}, \bibinfo{author}{H.~Lee},
  \bibinfo{author}{A.~Y. Ng},
\newblock \bibinfo{title}{Multimodal deep learning},
\newblock in: \bibinfo{booktitle}{Proceedings of the 28th international
  conference on machine learning (ICML-11)}, \bibinfo{year}{2011}, pp.
  \bibinfo{pages}{689--696}. \DOIprefix\doi{10.5555/3104482.3104569}.
%Type = Inproceedings
\bibitem[{He et~al.(2016)He, Zhang, Ren, and Sun}]{8}
\bibinfo{author}{K.~He}, \bibinfo{author}{X.~Zhang}, \bibinfo{author}{S.~Ren},
  \bibinfo{author}{J.~Sun},
\newblock \bibinfo{title}{Deep residual learning for image recognition},
\newblock in: \bibinfo{booktitle}{Proceedings of the IEEE conference on
  computer vision and pattern recognition}, \bibinfo{year}{2016}, pp.
  \bibinfo{pages}{770--778}. \DOIprefix\doi{10.1109/CVPR.2016.90}.
%Type = Inproceedings
\bibitem[{Huang et~al.(2017)Huang, Liu, Van Der~Maaten, and Weinberger}]{9}
\bibinfo{author}{G.~Huang}, \bibinfo{author}{Z.~Liu}, \bibinfo{author}{L.~Van
  Der~Maaten}, \bibinfo{author}{K.~Q. Weinberger},
\newblock \bibinfo{title}{Densely connected convolutional networks},
\newblock in: \bibinfo{booktitle}{Proceedings of the IEEE conference on
  computer vision and pattern recognition}, \bibinfo{year}{2017}, pp.
  \bibinfo{pages}{4700--4708}. \DOIprefix\doi{10.1109/CVPR.2017.243}.
%Type = Article
\bibitem[{Zhao et~al.(2023)Zhao, Zhou, Li, Tang, Wang, Hou, Min, Zhang, Zhang,
  Dong et~al.}]{10}
\bibinfo{author}{W.~X. Zhao}, \bibinfo{author}{K.~Zhou},
  \bibinfo{author}{J.~Li}, \bibinfo{author}{T.~Tang},
  \bibinfo{author}{X.~Wang}, \bibinfo{author}{Y.~Hou},
  \bibinfo{author}{Y.~Min}, \bibinfo{author}{B.~Zhang},
  \bibinfo{author}{J.~Zhang}, \bibinfo{author}{Z.~Dong}, et~al.,
\newblock \bibinfo{title}{A survey of large language models},
\newblock \bibinfo{journal}{arXiv preprint arXiv:2303.18223}
  (\bibinfo{year}{2023}). \DOIprefix\doi{10.48550/arXiv.2303.18223}.
%Type = Article
\bibitem[{Liu et~al.(2024)Liu, Zhang, Li, Yan, Gao, Chen, Yuan, Huang, Sun, Gao
  et~al.}]{11}
\bibinfo{author}{Y.~Liu}, \bibinfo{author}{K.~Zhang}, \bibinfo{author}{Y.~Li},
  \bibinfo{author}{Z.~Yan}, \bibinfo{author}{C.~Gao},
  \bibinfo{author}{R.~Chen}, \bibinfo{author}{Z.~Yuan},
  \bibinfo{author}{Y.~Huang}, \bibinfo{author}{H.~Sun},
  \bibinfo{author}{J.~Gao}, et~al.,
\newblock \bibinfo{title}{Sora: A review on background, technology,
  limitations, and opportunities of large vision models},
\newblock \bibinfo{journal}{arXiv preprint arXiv:2402.17177}
  (\bibinfo{year}{2024}). \DOIprefix\doi{10.48550/arXiv.2402.17177}.
%Type = Article
\bibitem[{Basheer and Hajmeer(2000)}]{12}
\bibinfo{author}{I.~A. Basheer}, \bibinfo{author}{M.~Hajmeer},
\newblock \bibinfo{title}{Artificial neural networks: fundamentals, computing,
  design, and application},
\newblock \bibinfo{journal}{Journal of microbiological methods}
  \bibinfo{volume}{43} (\bibinfo{year}{2000}) \bibinfo{pages}{3--31}.
  \DOIprefix\doi{10.1016/S0167-7012(00)00201-3}.
%Type = Article
\bibitem[{Laughlin and Sejnowski(2003)}]{13}
\bibinfo{author}{S.~B. Laughlin}, \bibinfo{author}{T.~J. Sejnowski},
\newblock \bibinfo{title}{Communication in neuronal networks},
\newblock \bibinfo{journal}{Science} \bibinfo{volume}{301}
  (\bibinfo{year}{2003}) \bibinfo{pages}{1870--1874}.
  \DOIprefix\doi{10.1126/science.1089662}.
%Type = Article
\bibitem[{Wang et~al.(2022)Wang, Pal, Yang, Kant, Zhu, and Guo}]{14}
\bibinfo{author}{J.~Wang}, \bibinfo{author}{A.~Pal}, \bibinfo{author}{Q.~Yang},
  \bibinfo{author}{K.~Kant}, \bibinfo{author}{K.~Zhu},
  \bibinfo{author}{S.~Guo},
\newblock \bibinfo{title}{Collaborative machine learning: Schemes, robustness,
  and privacy},
\newblock \bibinfo{journal}{IEEE Transactions on Neural Networks and Learning
  Systems}  (\bibinfo{year}{2022}). \DOIprefix\doi{10.1109/TNNLS.2022.3169347}.
%Type = Article
\bibitem[{Zhou et~al.(2022)Zhou, Liu, Qiao, Xiang, and Loy}]{15}
\bibinfo{author}{K.~Zhou}, \bibinfo{author}{Z.~Liu}, \bibinfo{author}{Y.~Qiao},
  \bibinfo{author}{T.~Xiang}, \bibinfo{author}{C.~C. Loy},
\newblock \bibinfo{title}{Domain generalization: A survey},
\newblock \bibinfo{journal}{IEEE Transactions on Pattern Analysis and Machine
  Intelligence}  (\bibinfo{year}{2022}).
  \DOIprefix\doi{10.1109/TPAMI.2022.3195549}.
%Type = Article
\bibitem[{Bassi et~al.(2024)Bassi, Dertkigil, and Cavalli}]{16}
\bibinfo{author}{P.~R. Bassi}, \bibinfo{author}{S.~S. Dertkigil},
  \bibinfo{author}{A.~Cavalli},
\newblock \bibinfo{title}{Improving deep neural network generalization and
  robustness to background bias via layer-wise relevance propagation
  optimization},
\newblock \bibinfo{journal}{Nature Communications} \bibinfo{volume}{15}
  (\bibinfo{year}{2024}) \bibinfo{pages}{291}.
  \DOIprefix\doi{10.1038/s41467-023-44371-z}.
%Type = Article
\bibitem[{Coles(2021)}]{17}
\bibinfo{author}{P.~J. Coles},
\newblock \bibinfo{title}{Seeking quantum advantage for neural networks},
\newblock \bibinfo{journal}{Nature Computational Science} \bibinfo{volume}{1}
  (\bibinfo{year}{2021}) \bibinfo{pages}{389--390}.
  \DOIprefix\doi{10.1038/s43588-021-00088-x}.
%Type = Inproceedings
\bibitem[{Ye et~al.(2020)Ye, Liu, Li, and Jiao}]{18}
\bibinfo{author}{W.~Ye}, \bibinfo{author}{R.~Liu}, \bibinfo{author}{Y.~Li},
  \bibinfo{author}{L.~Jiao},
\newblock \bibinfo{title}{Quantum-inspired evolutionary algorithm for
  convolutional neural networks architecture search},
\newblock in: \bibinfo{booktitle}{2020 IEEE Congress on Evolutionary
  Computation (CEC)}, \bibinfo{organization}{IEEE}, \bibinfo{year}{2020}, pp.
  \bibinfo{pages}{1--8}. \DOIprefix\doi{10.1109/CEC48606.2020.9185727}.
%Type = Inproceedings
\bibitem[{Song et~al.(2021)Song, Hou, and Liu}]{19}
\bibinfo{author}{S.~Song}, \bibinfo{author}{Y.~Hou}, \bibinfo{author}{G.~Liu},
\newblock \bibinfo{title}{The interpretability of quantum-inspired neural
  network},
\newblock in: \bibinfo{booktitle}{2021 4th International Conference on
  Artificial Intelligence and Big Data (ICAIBD)}, \bibinfo{organization}{IEEE},
  \bibinfo{year}{2021}, pp. \bibinfo{pages}{294--298}.
  \DOIprefix\doi{10.1109/ICAIBD51990.2021.9459009}.
%Type = Article
\bibitem[{Wang et~al.(2022)Wang, Lin, and Wu}]{20}
\bibinfo{author}{L.-J. Wang}, \bibinfo{author}{J.-Y. Lin},
  \bibinfo{author}{S.~Wu},
\newblock \bibinfo{title}{Implementation of quantum stochastic walks for
  function approximation, two-dimensional data classification, and sequence
  classification},
\newblock \bibinfo{journal}{Physical Review Research} \bibinfo{volume}{4}
  (\bibinfo{year}{2022}) \bibinfo{pages}{023058}.
  \DOIprefix\doi{10.1103/PhysRevResearch.4.023058}.
%Type = Inproceedings
\bibitem[{Szwarcman et~al.(2019)Szwarcman, Civitarese, and Vellasco}]{21}
\bibinfo{author}{D.~Szwarcman}, \bibinfo{author}{D.~Civitarese},
  \bibinfo{author}{M.~Vellasco},
\newblock \bibinfo{title}{Quantum-inspired neural architecture search},
\newblock in: \bibinfo{booktitle}{2019 International Joint Conference on Neural
  Networks (IJCNN)}, \bibinfo{organization}{IEEE}, \bibinfo{year}{2019}, pp.
  \bibinfo{pages}{1--8}. \DOIprefix\doi{10.1109/IJCNN.2019.8852453}.
%Type = Article
\bibitem[{Mitarai et~al.(2018)Mitarai, Negoro, Kitagawa, and Fujii}]{22}
\bibinfo{author}{K.~Mitarai}, \bibinfo{author}{M.~Negoro},
  \bibinfo{author}{M.~Kitagawa}, \bibinfo{author}{K.~Fujii},
\newblock \bibinfo{title}{Quantum circuit learning},
\newblock \bibinfo{journal}{Physical Review A} \bibinfo{volume}{98}
  (\bibinfo{year}{2018}) \bibinfo{pages}{032309}.
  \DOIprefix\doi{10.1103/PhysRevA.98.032309}.
%Type = Article
\bibitem[{Liang et~al.(2021)Liang, Peng, Zheng, Silv{\'e}n, and Zhao}]{23}
\bibinfo{author}{Y.~Liang}, \bibinfo{author}{W.~Peng}, \bibinfo{author}{Z.-J.
  Zheng}, \bibinfo{author}{O.~Silv{\'e}n}, \bibinfo{author}{G.~Zhao},
\newblock \bibinfo{title}{A hybrid quantum--classical neural network with deep
  residual learning},
\newblock \bibinfo{journal}{Neural Networks} \bibinfo{volume}{143}
  (\bibinfo{year}{2021}) \bibinfo{pages}{133--147}.
  \DOIprefix\doi{10.1016/j.neunet.2021.05.028}.
%Type = Article
\bibitem[{Schetakis et~al.(2022)Schetakis, Aghamalyan, Griffin, and
  Boguslavsky}]{24}
\bibinfo{author}{N.~Schetakis}, \bibinfo{author}{D.~Aghamalyan},
  \bibinfo{author}{P.~Griffin}, \bibinfo{author}{M.~Boguslavsky},
\newblock \bibinfo{title}{Review of some existing qml frameworks and novel
  hybrid classical--quantum neural networks realising binary classification for
  the noisy datasets},
\newblock \bibinfo{journal}{Scientific Reports} \bibinfo{volume}{12}
  (\bibinfo{year}{2022}) \bibinfo{pages}{11927}.
  \DOIprefix\doi{10.1038/s41598-022-14876-6}.
%Type = Article
\bibitem[{Konar et~al.(2023)Konar, Sarma, Bhandary, Bhattacharyya, Cangi, and
  Aggarwal}]{25}
\bibinfo{author}{D.~Konar}, \bibinfo{author}{A.~D. Sarma},
  \bibinfo{author}{S.~Bhandary}, \bibinfo{author}{S.~Bhattacharyya},
  \bibinfo{author}{A.~Cangi}, \bibinfo{author}{V.~Aggarwal},
\newblock \bibinfo{title}{A shallow hybrid classical--quantum spiking
  feedforward neural network for noise-robust image classification},
\newblock \bibinfo{journal}{Applied Soft Computing} \bibinfo{volume}{136}
  (\bibinfo{year}{2023}) \bibinfo{pages}{110099}.
  \DOIprefix\doi{10.1016/j.asoc.2023.110099}.
%Type = Article
\bibitem[{Li et~al.(2013)Li, Xiao, Shang, Tong, Li, and Cao}]{26}
\bibinfo{author}{P.~Li}, \bibinfo{author}{H.~Xiao}, \bibinfo{author}{F.~Shang},
  \bibinfo{author}{X.~Tong}, \bibinfo{author}{X.~Li}, \bibinfo{author}{M.~Cao},
\newblock \bibinfo{title}{A hybrid quantum-inspired neural networks with
  sequence inputs},
\newblock \bibinfo{journal}{Neurocomputing} \bibinfo{volume}{117}
  (\bibinfo{year}{2013}) \bibinfo{pages}{81--90}.
  \DOIprefix\doi{10.1016/j.neucom.2013.01.029}.
%Type = Article
\bibitem[{Wang et~al.(2025)Wang, Mao, Li, Li, and Li}]{49}
\bibinfo{author}{A.~Wang}, \bibinfo{author}{D.~Mao}, \bibinfo{author}{X.~Li},
  \bibinfo{author}{T.~Li}, \bibinfo{author}{L.~Li},
\newblock \bibinfo{title}{Hqnet: A hybrid quantum network for multi-class mri
  brain classification via quantum computing},
\newblock \bibinfo{journal}{Expert Systems with Applications}
  \bibinfo{volume}{261} (\bibinfo{year}{2025}) \bibinfo{pages}{125537}.
  \URLprefix
  \url{https://www.sciencedirect.com/science/article/pii/S0957417424024047}.
  \DOIprefix\doi{https://doi.org/10.1016/j.eswa.2024.125537}.
%Type = Article
\bibitem[{Hong et~al.(2024)Hong, Rioflorido, and Zhang}]{50}
\bibinfo{author}{Y.-Y. Hong}, \bibinfo{author}{C.~L. P.~P. Rioflorido},
  \bibinfo{author}{W.~Zhang},
\newblock \bibinfo{title}{Hybrid deep learning and quantum-inspired neural
  network for day-ahead spatiotemporal wind speed forecasting},
\newblock \bibinfo{journal}{Expert Systems with Applications}
  \bibinfo{volume}{241} (\bibinfo{year}{2024}) \bibinfo{pages}{122645}.
  \URLprefix
  \url{https://www.sciencedirect.com/science/article/pii/S0957417423031470}.
  \DOIprefix\doi{https://doi.org/10.1016/j.eswa.2023.122645}.
%Type = Article
\bibitem[{Sagingalieva et~al.(2023)Sagingalieva, Kordzanganeh, Kurkin,
  Melnikov, Kuhmistrov, Perelshtein, Melnikov, Skolik, and Dollen}]{51}
\bibinfo{author}{A.~Sagingalieva}, \bibinfo{author}{M.~Kordzanganeh},
  \bibinfo{author}{A.~Kurkin}, \bibinfo{author}{A.~Melnikov},
  \bibinfo{author}{D.~Kuhmistrov}, \bibinfo{author}{M.~Perelshtein},
  \bibinfo{author}{A.~Melnikov}, \bibinfo{author}{A.~Skolik},
  \bibinfo{author}{D.~V. Dollen},
\newblock \bibinfo{title}{Hybrid quantum resnet for car classification and its
  hyperparameter optimization},
\newblock \bibinfo{journal}{Quantum Machine Intelligence} \bibinfo{volume}{5}
  (\bibinfo{year}{2023}). \DOIprefix\doi{10.1007/s42484-023-00123-2}.
%Type = Article
\bibitem[{Clements et~al.(2016)Clements, Humphreys, Metcalf, Kolthammer, and
  Walmsley}]{27}
\bibinfo{author}{W.~R. Clements}, \bibinfo{author}{P.~C. Humphreys},
  \bibinfo{author}{B.~J. Metcalf}, \bibinfo{author}{W.~S. Kolthammer},
  \bibinfo{author}{I.~A. Walmsley},
\newblock \bibinfo{title}{Optimal design for universal multiport
  interferometers},
\newblock \bibinfo{journal}{Optica} \bibinfo{volume}{3} (\bibinfo{year}{2016})
  \bibinfo{pages}{1460--1465}. \DOIprefix\doi{10.1364/OPTICA.3.001460}.
%Type = Article
\bibitem[{Pedregosa et~al.(2011)Pedregosa, Varoquaux, Gramfort, Michel,
  Thirion, Grisel, Blondel, Prettenhofer, Weiss, Dubourg et~al.}]{28}
\bibinfo{author}{F.~Pedregosa}, \bibinfo{author}{G.~Varoquaux},
  \bibinfo{author}{A.~Gramfort}, \bibinfo{author}{V.~Michel},
  \bibinfo{author}{B.~Thirion}, \bibinfo{author}{O.~Grisel},
  \bibinfo{author}{M.~Blondel}, \bibinfo{author}{P.~Prettenhofer},
  \bibinfo{author}{R.~Weiss}, \bibinfo{author}{V.~Dubourg}, et~al.,
\newblock \bibinfo{title}{Scikit-learn: Machine learning in python},
\newblock \bibinfo{journal}{the Journal of machine Learning research}
  \bibinfo{volume}{12} (\bibinfo{year}{2011}) \bibinfo{pages}{2825--2830}.
  \URLprefix \url{http://jmlr.org/papers/v12/pedregosa11a.html}.
%Type = Inproceedings
\bibitem[{Krizhevsky et~al.(2012)Krizhevsky, Sutskever, and Hinton}]{29}
\bibinfo{author}{A.~Krizhevsky}, \bibinfo{author}{I.~Sutskever},
  \bibinfo{author}{G.~E. Hinton},
\newblock \bibinfo{title}{Imagenet classification with deep convolutional
  neural networks},
\newblock in: \bibinfo{editor}{F.~Pereira}, \bibinfo{editor}{C.~Burges},
  \bibinfo{editor}{L.~Bottou}, \bibinfo{editor}{K.~Weinberger} (Eds.),
  \bibinfo{booktitle}{Advances in Neural Information Processing Systems},
  volume~\bibinfo{volume}{25}, \bibinfo{publisher}{Curran Associates, Inc.},
  \bibinfo{year}{2012}, pp. \bibinfo{pages}{84--90}.
  \DOIprefix\doi{10.1145/3065386}.
%Type = Article
\bibitem[{Ren et~al.(2016)Ren, He, Girshick, and Sun}]{30}
\bibinfo{author}{S.~Ren}, \bibinfo{author}{K.~He},
  \bibinfo{author}{R.~Girshick}, \bibinfo{author}{J.~Sun},
\newblock \bibinfo{title}{Faster r-cnn: Towards real-time object detection with
  region proposal networks},
\newblock \bibinfo{journal}{IEEE transactions on pattern analysis and machine
  intelligence} \bibinfo{volume}{39} (\bibinfo{year}{2016})
  \bibinfo{pages}{1137--1149}. \DOIprefix\doi{10.1109/TPAMI.2016.2577031}.
%Type = Inproceedings
\bibitem[{Chen et~al.(2021)Chen, Wang, Yang, Zhang, Cheng, and Sun}]{31}
\bibinfo{author}{Q.~Chen}, \bibinfo{author}{Y.~Wang},
  \bibinfo{author}{T.~Yang}, \bibinfo{author}{X.~Zhang},
  \bibinfo{author}{J.~Cheng}, \bibinfo{author}{J.~Sun},
\newblock \bibinfo{title}{You only look one-level feature},
\newblock in: \bibinfo{booktitle}{Proceedings of the IEEE/CVF conference on
  computer vision and pattern recognition}, \bibinfo{year}{2021}, pp.
  \bibinfo{pages}{13039--13048}. \DOIprefix\doi{10.1109/CVPR46437.2021.01284}.
%Type = Inproceedings
\bibitem[{Lin et~al.(2021)Lin, Koprinska, and Rana}]{32}
\bibinfo{author}{Y.~Lin}, \bibinfo{author}{I.~Koprinska},
  \bibinfo{author}{M.~Rana},
\newblock \bibinfo{title}{Ssdnet: State space decomposition neural network for
  time series forecasting},
\newblock in: \bibinfo{booktitle}{2021 IEEE International Conference on Data
  Mining (ICDM)}, \bibinfo{organization}{IEEE}, \bibinfo{year}{2021}, pp.
  \bibinfo{pages}{370--378}. \DOIprefix\doi{10.1109/ICDM51629.2021.00048}.
%Type = Inproceedings
\bibitem[{Zhao et~al.(2017)Zhao, Shi, Qi, Wang, and Jia}]{33}
\bibinfo{author}{H.~Zhao}, \bibinfo{author}{J.~Shi}, \bibinfo{author}{X.~Qi},
  \bibinfo{author}{X.~Wang}, \bibinfo{author}{J.~Jia},
\newblock \bibinfo{title}{Pyramid scene parsing network},
\newblock in: \bibinfo{booktitle}{Proceedings of the IEEE conference on
  computer vision and pattern recognition}, \bibinfo{year}{2017}, pp.
  \bibinfo{pages}{2881--2890}. \DOIprefix\doi{10.1109/CVPR.2017.660}.
%Type = Inproceedings
\bibitem[{Deng et~al.(2009)Deng, Dong, Socher, Li, Li, and Fei-Fei}]{34}
\bibinfo{author}{J.~Deng}, \bibinfo{author}{W.~Dong},
  \bibinfo{author}{R.~Socher}, \bibinfo{author}{L.-J. Li},
  \bibinfo{author}{K.~Li}, \bibinfo{author}{L.~Fei-Fei},
\newblock \bibinfo{title}{Imagenet: A large-scale hierarchical image database},
\newblock in: \bibinfo{booktitle}{2009 IEEE conference on computer vision and
  pattern recognition}, \bibinfo{organization}{Ieee}, \bibinfo{year}{2009}, pp.
  \bibinfo{pages}{248--255}. \DOIprefix\doi{10.1109/CVPR.2009.5206848}.
%Type = Article
\bibitem[{Minaee et~al.(2021)Minaee, Boykov, Porikli, Plaza, Kehtarnavaz, and
  Terzopoulos}]{35}
\bibinfo{author}{S.~Minaee}, \bibinfo{author}{Y.~Boykov},
  \bibinfo{author}{F.~Porikli}, \bibinfo{author}{A.~Plaza},
  \bibinfo{author}{N.~Kehtarnavaz}, \bibinfo{author}{D.~Terzopoulos},
\newblock \bibinfo{title}{Image segmentation using deep learning: A survey},
\newblock \bibinfo{journal}{IEEE transactions on pattern analysis and machine
  intelligence} \bibinfo{volume}{44} (\bibinfo{year}{2021})
  \bibinfo{pages}{3523--3542}. \DOIprefix\doi{10.1109/TPAMI.2021.3059968}.
%Type = Article
\bibitem[{Hering et~al.(2022)Hering, Hansen, Mok, Chung, Siebert, H{\"a}ger,
  Lange, Kuckertz, Heldmann, Shao et~al.}]{36}
\bibinfo{author}{A.~Hering}, \bibinfo{author}{L.~Hansen},
  \bibinfo{author}{T.~C. Mok}, \bibinfo{author}{A.~C. Chung},
  \bibinfo{author}{H.~Siebert}, \bibinfo{author}{S.~H{\"a}ger},
  \bibinfo{author}{A.~Lange}, \bibinfo{author}{S.~Kuckertz},
  \bibinfo{author}{S.~Heldmann}, \bibinfo{author}{W.~Shao}, et~al.,
\newblock \bibinfo{title}{Learn2reg: comprehensive multi-task medical image
  registration challenge, dataset and evaluation in the era of deep learning},
\newblock \bibinfo{journal}{IEEE Transactions on Medical Imaging}
  \bibinfo{volume}{42} (\bibinfo{year}{2022}) \bibinfo{pages}{697--712}.
  \DOIprefix\doi{10.1109/TMI.2022.3213983}.
%Type = Article
\bibitem[{Li et~al.(2021)Li, Liu, Yang, Peng, and Zhou}]{37}
\bibinfo{author}{Z.~Li}, \bibinfo{author}{F.~Liu}, \bibinfo{author}{W.~Yang},
  \bibinfo{author}{S.~Peng}, \bibinfo{author}{J.~Zhou},
\newblock \bibinfo{title}{A survey of convolutional neural networks: analysis,
  applications, and prospects},
\newblock \bibinfo{journal}{IEEE transactions on neural networks and learning
  systems} \bibinfo{volume}{33} (\bibinfo{year}{2021})
  \bibinfo{pages}{6999--7019}. \DOIprefix\doi{10.1109/TNNLS.2021.3084827}.
%Type = Article
\bibitem[{Vaswani et~al.(2017)Vaswani, Shazeer, Parmar, Uszkoreit, Jones,
  Gomez, Kaiser, and Polosukhin}]{38}
\bibinfo{author}{A.~Vaswani}, \bibinfo{author}{N.~Shazeer},
  \bibinfo{author}{N.~Parmar}, \bibinfo{author}{J.~Uszkoreit},
  \bibinfo{author}{L.~Jones}, \bibinfo{author}{A.~N. Gomez},
  \bibinfo{author}{{\L}.~Kaiser}, \bibinfo{author}{I.~Polosukhin},
\newblock \bibinfo{title}{Attention is all you need},
\newblock \bibinfo{journal}{Advances in neural information processing systems}
  \bibinfo{volume}{30} (\bibinfo{year}{2017}) \bibinfo{pages}{6000--6010}.
  \DOIprefix\doi{10.5555/3295222.3295349}.
%Type = Article
\bibitem[{Chiribella et~al.(2008)Chiribella, D’Ariano, and Perinotti}]{39}
\bibinfo{author}{G.~Chiribella}, \bibinfo{author}{G.~M. D’Ariano},
  \bibinfo{author}{P.~Perinotti},
\newblock \bibinfo{title}{Quantum circuit architecture},
\newblock \bibinfo{journal}{Physical review letters} \bibinfo{volume}{101}
  (\bibinfo{year}{2008}) \bibinfo{pages}{060401}.
  \DOIprefix\doi{10.1103/PhysRevLett.101.060401}.
%Type = Inproceedings
\bibitem[{Milletari et~al.(2016)Milletari, Navab, and Ahmadi}]{40}
\bibinfo{author}{F.~Milletari}, \bibinfo{author}{N.~Navab},
  \bibinfo{author}{S.-A. Ahmadi},
\newblock \bibinfo{title}{V-net: Fully convolutional neural networks for
  volumetric medical image segmentation},
\newblock in: \bibinfo{booktitle}{2016 fourth international conference on 3D
  vision (3DV)}, \bibinfo{organization}{Ieee}, \bibinfo{year}{2016}, pp.
  \bibinfo{pages}{565--571}. \DOIprefix\doi{10.1109/3DV.2016.79}.
%Type = Book
\bibitem[{Nielsen and Chuang(2010)}]{nie}
\bibinfo{author}{M.~A. Nielsen}, \bibinfo{author}{I.~L. Chuang},
  \bibinfo{title}{Quantum computation and quantum information},
  \bibinfo{publisher}{Cambridge university press}, \bibinfo{year}{2010}.
%Type = Book
\bibitem[{Bies and Howard(2017)}]{41}
\bibinfo{author}{C.~H.~H. Bies, David~A.}, \bibinfo{author}{C.~Q. Howard},
  \bibinfo{title}{Engineering Noise Control}, \bibinfo{publisher}{CRC Press},
  \bibinfo{year}{2017}. \DOIprefix\doi{10.1201/9781351228152}.
%Type = Inproceedings
\bibitem[{Schmidt and Lipson(2007)}]{42}
\bibinfo{author}{M.~D. Schmidt}, \bibinfo{author}{H.~Lipson},
\newblock \bibinfo{title}{Learning noise},
\newblock in: \bibinfo{booktitle}{Proceedings of the 9th annual conference on
  Genetic and evolutionary computation}, \bibinfo{year}{2007}, pp.
  \bibinfo{pages}{1680--1685}. \DOIprefix\doi{10.1145/1276958.1277289}.
%Type = Article
\bibitem[{Kosko et~al.(2020)Kosko, Audhkhasi, and Osoba}]{43}
\bibinfo{author}{B.~Kosko}, \bibinfo{author}{K.~Audhkhasi},
  \bibinfo{author}{O.~Osoba},
\newblock \bibinfo{title}{Noise can speed backpropagation learning and deep
  bidirectional pretraining},
\newblock \bibinfo{journal}{Neural Networks} \bibinfo{volume}{129}
  (\bibinfo{year}{2020}) \bibinfo{pages}{359--384}.
  \DOIprefix\doi{10.1016/j.neunet.2020.04.004}.
%Type = Article
\bibitem[{Semenova et~al.(2022)Semenova, Larger, and Brunner}]{44}
\bibinfo{author}{N.~Semenova}, \bibinfo{author}{L.~Larger},
  \bibinfo{author}{D.~Brunner},
\newblock \bibinfo{title}{Understanding and mitigating noise in trained deep
  neural networks},
\newblock \bibinfo{journal}{Neural Networks} \bibinfo{volume}{146}
  (\bibinfo{year}{2022}) \bibinfo{pages}{151--160}.
  \DOIprefix\doi{10.1016/j.neunet.2021.11.008}.
%Type = Article
\bibitem[{Xiao et~al.(2024)Xiao, Adegok, Leung, and Leung}]{45}
\bibinfo{author}{Y.~Xiao}, \bibinfo{author}{M.~Adegok}, \bibinfo{author}{C.-S.
  Leung}, \bibinfo{author}{K.~W. Leung},
\newblock \bibinfo{title}{Robust noise-aware algorithm for randomized neural
  network and its convergence properties},
\newblock \bibinfo{journal}{Neural Networks}  (\bibinfo{year}{2024})
  \bibinfo{pages}{106202}. \DOIprefix\doi{10.1016/j.neunet.2024.106202}.
%Type = Article
\bibitem[{Shen and Wang(2011)}]{46}
\bibinfo{author}{Y.~Shen}, \bibinfo{author}{J.~Wang},
\newblock \bibinfo{title}{Robustness analysis of global exponential stability
  of recurrent neural networks in the presence of time delays and random
  disturbances},
\newblock \bibinfo{journal}{IEEE transactions on neural networks and learning
  systems} \bibinfo{volume}{23} (\bibinfo{year}{2011}) \bibinfo{pages}{87--96}.
  \DOIprefix\doi{10.1109/TNNLS.2011.2178326}.
%Type = Article
\bibitem[{Nettleton et~al.(2010)Nettleton, Orriols-Puig, and Fornells}]{47}
\bibinfo{author}{D.~F. Nettleton}, \bibinfo{author}{A.~Orriols-Puig},
  \bibinfo{author}{A.~Fornells},
\newblock \bibinfo{title}{A study of the effect of different types of noise on
  the precision of supervised learning techniques},
\newblock \bibinfo{journal}{Artificial intelligence review}
  \bibinfo{volume}{33} (\bibinfo{year}{2010}) \bibinfo{pages}{275--306}.
  \DOIprefix\doi{10.1007/s10462-010-9156-z}.
%Type = Article
\bibitem[{Lu et~al.(2017)Lu, Pu, Wang, Hu, and Wang}]{48}
\bibinfo{author}{Z.~Lu}, \bibinfo{author}{H.~Pu}, \bibinfo{author}{F.~Wang},
  \bibinfo{author}{Z.~Hu}, \bibinfo{author}{L.~Wang},
\newblock \bibinfo{title}{The expressive power of neural networks: A view from
  the width},
\newblock \bibinfo{journal}{Advances in neural information processing systems}
  \bibinfo{volume}{30} (\bibinfo{year}{2017}).
  \DOIprefix\doi{10.5555/3295222.3295371}.

\end{thebibliography}

\end{document}